\documentclass{article}
\usepackage{fullpage}
\usepackage[svgnames,dvipsnames,color,table]{xcolor}
\usepackage{hyperref}
\hypersetup{colorlinks,linkcolor={blue},citecolor={blue},urlcolor={RoyalBlue}}

\usepackage{amsthm}
\usepackage{amsmath} %
\usepackage{nicefrac}
\usepackage{bm}
\usepackage{booktabs}
\usepackage{amsfonts}
\usepackage{bbm}
\usepackage{mleftright}
\usepackage{dsfont}
\usepackage{amscd,amssymb,amsmath,amsthm,bbold}
\usepackage{graphicx}
\usepackage{float}
\usepackage{multirow}
\usepackage[numbers]{natbib}

\usepackage{pifont}
\newcommand{\cmark}{\ding{51}}%
\usepackage{microtype}
\usepackage{enumitem}
\usepackage{xfrac}
\usepackage{parskip}
\usepackage{numprint}
\usepackage{booktabs} %

\usepackage{dblfloatfix}
\usepackage{transparent}

\usepackage{algorithm}
\usepackage{listings}
\usepackage{titling}

\usepackage[textsize=scriptsize,textwidth=1.9cm]{todonotes}
\usepackage{xspace}
\usepackage[margin=1cm]{caption}
\usepackage{subcaption}
\usepackage{etoolbox}
\usepackage{comment}
\usepackage{etoc}
\usepackage{wrapfig}
\usepackage{placeins}
\usepackage{makecell}
\usepackage{lipsum,xcolor}

\newcommand{\customfootnotetext}[2]{{%
  \renewcommand{\thefootnote}{#1}%
  \footnotetext[0]{#2}}}%

\usepackage{cleveref}
\crefformat{section}{\S#2#1#3} %
\crefformat{subsection}{\S#2#1#3}
\crefformat{subsubsection}{\S#2#1#3}

\usepackage{adjustbox}
\usepackage{array}

\newcolumntype{R}[2]{%
    >{\adjustbox{angle=#1,lap=\width-(#2)}\bgroup}%
    l%
    <{\egroup}%
}

\begin{document}

\date{}
\title{Data Determines Distributional Robustness \\ in Contrastive Language Image Pre-training (CLIP)}
\author{
        Alex Fang$^{\dag}$ \\
        \and
        Gabriel Ilharco$^{\dag}$\\
        \and
        Mitchell Wortsman$^{\dag}$\\
        \and
        Yuhao Wan$^{\dag}$\\
        \and
        Vaishaal Shankar$^\diamond$ \\
        \and
        Achal Dave$^\diamond$ \\
        \and
        Ludwig Schmidt$^{\dag\circ}$ \\ %
}

\setlength{\droptitle}{-0.5cm}
\maketitle

\customfootnotetext{$\dag$}
{
University of Washington, \{apf1, gamaga, mitchnw, yuhaowan, schmidt\}@cs.washington.edu

$^\diamond$Amazon, \{vaishaal, achald\}@amazon.com

$^\circ$Allen Institute for Artificial Intelligence
}

\newif\ifcomments
\commentsfalse

\ifcomments
    \newcommand\achal[1]{{\todo[color=teal]{AD: {#1}}}}
    \newcommand\alex[1]{{\todo[color=olive]{AF: {#1}}}}
    \newcommand\gamaga[1]{{\todo[color=yellow]{GI: {#1}}}}
    \newcommand\ludwig[1]{{\todo[color=orange!50]{LS: {#1}}}}
    \newcommand\mitchell[1]{{\todo[color=purple]{MW: {#1}}}}
    \newcommand\vaishaal[1]{{\todo[color=green]{VS: {#1}}}}
    \newcommand\yuhao[1]{{\todo[color=blue!20]{YW: {#1}}}}
\else
    \providecommand{\achal}[1]{}
    \providecommand{\alex}[1]{}
    \providecommand{\gamaga}[1]{}
    \providecommand{\mitchell}[1]{}
    \providecommand{\vaishaal}[1]{}
    \providecommand{\ludwig}[1]{}
    \providecommand{\yuhao}[1]{}
\fi

\newcommand{\numimagenetfallimages}{14197122}
\newcommand{\imagenetcaptions}{ImageNet-Captions}
\newcommand{\yfcccls}{YFCC-15M-Cls}
\newcommand{\baseline}{NoCLIP}

\makeatletter
\renewcommand\paragraph{\@startsection{paragraph}{4}{\z@}                                     {1.05ex \@plus1ex \@minus.2ex}                                {-.5em}
{\normalfont\normalsize\bfseries}}
\makeatother
\newcommand{\smallpara}[1]{\textbf{#1}}
\newcommand{\beforesec}{\vspace*{0em}}
\newcommand{\postsec}{\vspace*{0em}}

\vspace*{-1cm}
\begin{abstract}
Contrastively trained language-image models such as CLIP, ALIGN, and BASIC have demonstrated unprecedented robustness to multiple challenging natural distribution shifts.
Since these language-image models differ from previous training approaches in several ways, an important question is what causes the large robustness gains.
We answer this question via a systematic experimental investigation.
Concretely, we study five different possible causes for the robustness gains:
(i) the training set size,
(ii) the training distribution,
(iii) language supervision at training time,
(iv) language supervision at test time, and
(v) the contrastive loss function.
Our experiments show that the more diverse training distribution is the main cause for the robustness gains, with the other factors contributing little to no robustness.
Beyond our experimental results, we also introduce ImageNet-Captions, a version of ImageNet with original text annotations from Flickr, to enable further controlled experiments of language-image training.
\end{abstract}

\setcounter{footnote}{0}
\section{Introduction}
Large pre-trained language-image models such as CLIP \cite{radford21a}, ALIGN \cite{jia2021scaling}, and BASIC \cite{basic2021} have recently demonstrated unprecedented robustness on a variety of natural distribution shifts.
In contrast to prior models that are trained on images with class annotations, CLIP and relatives\footnote{Following \citet{radford21a}, we use \emph{CLIP} as a name for the general training technique, not only their specific models.} are directly trained on images and their corresponding unstructured text from the web.
The resulting models achieve large robustness even on challenging distribution shifts such as ImageNetV2 \cite{recht2019imagenet} and ObjectNet \cite{objectnet}.
No prior algorithmic techniques had enhanced robustness on these datasets even after multiple years of intensive research in reliable machine learning \cite{Djolonga_2021_CVPR,taori2020measuring}.
As CLIP also improves robustness on a wide range of other distribution shifts, an important question emerges: \emph{What causes CLIP’s unprecedented robustness?}

The fact that language-image models were the first to achieve large robustness gains suggests that multimodal learning on language and image data may be key to more robust image representations.
However, pinpointing the exact cause of CLIP's robustness is complicated by the fact that CLIP relied on several changes to the common supervised training paradigm for image classification models.
For instance, the CLIP models with highest accuracy follow the vision transformer (ViT) architecture \cite{vit2020}.
\citet{radford21a} already investigated model architecture and size, showing that these factors do not affect the robustness of their CLIP models.
Nevertheless, there is still a long list of possible causes for CLIP's robustness:
\begin{itemize}[itemsep=0pt,topsep=-3pt,parsep=1pt]
\item The large training set size (400 million images)
\item The training distribution
\item Language supervision at training time
\item Language supervision at test time via prompts
\item The contrastive loss function
\end{itemize}
Understanding the mechanism underlying CLIP’s robustness is important as it may guide the way towards more reliable machine learning more broadly.

In this paper, we answer the question of CLIP’s robustness via a series of controlled experiments that test the five possible causes listed above.
Our main result is that CLIP’s robustness is determined almost exclusively by the training distribution.
Language supervision at training time does \emph{not} make the resulting models more robust than standard supervised learning when the images in the training set are the same.
Hence language supervision only has an \emph{indirect} effect on robustness.
In particular, language supervision simplifies training on a diverse distribution of images by removing the need for consistent annotation with class labels.
The more diverse training distribution –– not the language supervision –– then leads to more robust representations.

\begin{figure}[ht]
    \centering
    \includegraphics[width=0.7\textwidth]{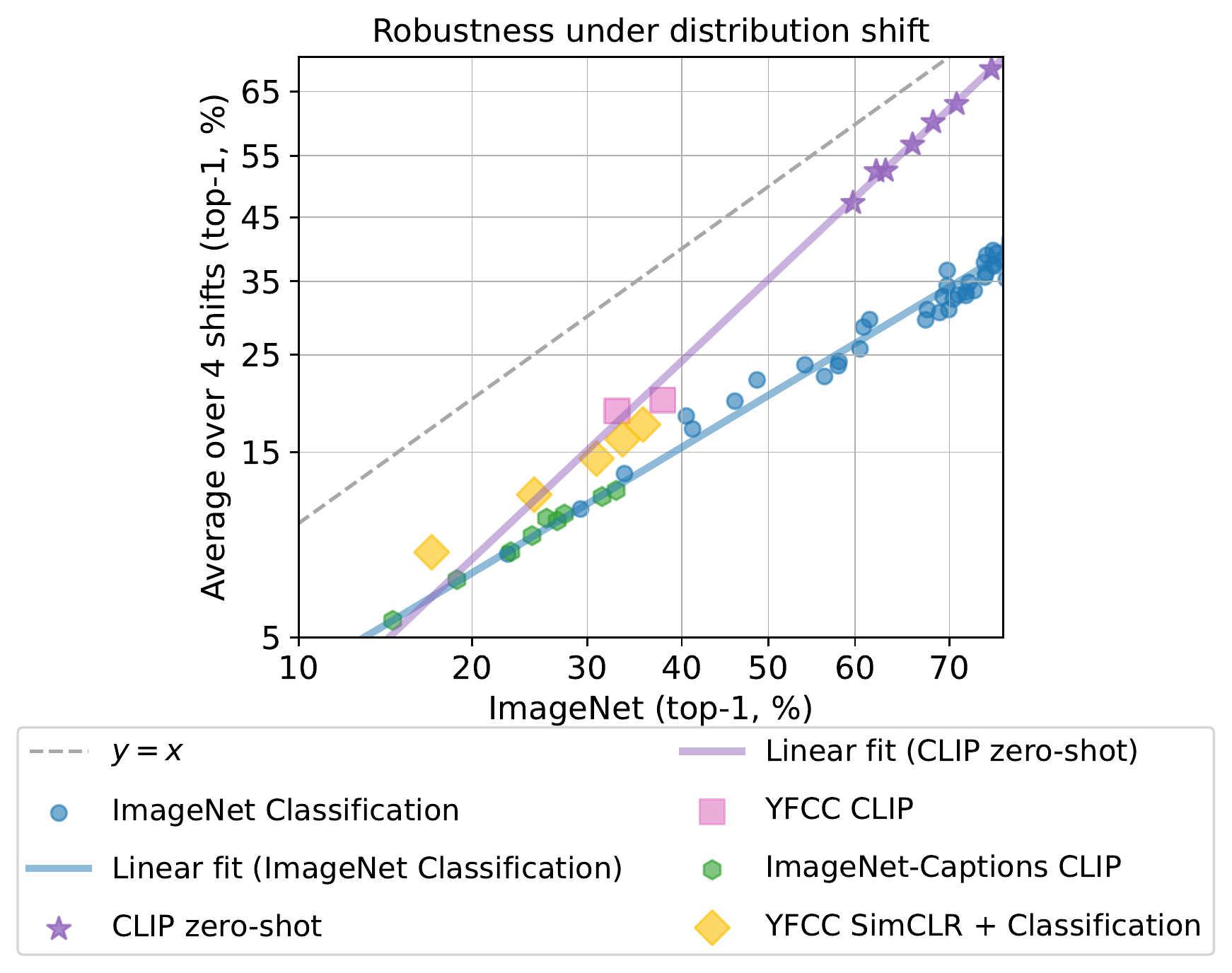}
    \caption{\label{fig:figure1} We compare models trained using different methods and on different datasets, measuring their robustness on a range of natural distribution shifts (ImageNetV2, ImageNet-R, ImageNet-Sketch, and ObjectNet).
    The CLIP models stand out with their consistent performance in the presence of distribution shift.
    We find that large gains in effective robustness (improvement over ImageNet models) only come from varying the training distribution.
    Language supervision alone does not cause robustness.}
\end{figure}

Our investigation of CLIP’s robustness rests on two further contributions.
First, we introduce ImageNet-Captions, a new dataset for training on paired language-image data.
ImageNet-Captions augments 463,622 of the 1.2 million images in the ImageNet 2012 training set \cite{russakovsky2015imagenet} with the original text data sourced from the corresponding Flickr images.
ImageNet-Captions enables controlled experiments comparing standard (class-based) ImageNet training with language-image training on the same set of images.
Such experiments precisely pinpoint the effect of utilizing language when training computer vision models.

Second, we provide a new baseline for language-image training that minimizes the interaction between the vision and language components yet achieves accuracy similar to CLIP training.
Specifically, we introduce the following training procedure and illustrate its behavior on the YFCC-15M dataset \cite{thomee2016yfcc100m,radford21a}:
\begin{enumerate}
\item Use SimCLR \cite{simclr2020} to pre-train a representation on only the \emph{images} in YFCC-15M.
\item Fine-tune the resulting representation by matching examples in YFCC-15M to ImageNet classes with simple \emph{text matches} in the corresponding captions.
\end{enumerate}
In particular, our approach relies on \emph{no} language model, demonstrating that it is possible to match the performance of CLIP training with much simpler language processing.
Besides serving as a useful baseline to understand CLIP training, our simplified approach may open the way for further algorithmic innovations in language-image training.

The remainder of our paper proceeds as follows.
The next two sections introduce relevant background and our new dataset ImageNet-Captions as experimental framework.
Sections~\ref{sec:experiments} and \ref{sec:yfcc} then describe our main experiments testing the impact of language supervision and the training distribution on the robustness of the resulting models.
Sections~\ref{sec:testprompts} and \ref{sec:loss} present the evidence against test time prompts and contrastive training losses as causes for CLIP’s robustness.
We summarize our findings in Section~\ref{sec:conclusion}.

\section{Background}
Pinpointing the cause of CLIP's robustness requires a precise experimental setup for comparing robustness across a range of models.
To this end, we follow the \emph{effective robustness} framework first proposed by \citet{taori2020measuring} and later utilized by \citet{radford21a} to demonstrate the robustness gains of their CLIP models.
We first review this measurement framework and then survey further related work.

\subsection{Experimental setup for measuring robustness}
An important goal of reliable machine learning is to design models that consistently perform well across a diverse range of test distributions.
For instance, a model that achieves 75\% accuracy on ImageNet should ideally also achieve 75\% accuracy on the closely related ImageNetV2 distribution shift because humans can do so \cite{shankar2020evaluating}.
But instead of consistent performance, most current ImageNet models see a 12 percentage point (pp) accuracy drop on this distribution shift \citep{recht2019imagenet}.
In contrast, the CLIP models of \citet{radford21a} are more robust and only have a 6 pp accuracy drop on ImageNetV2.
Compared to earlier models, CLIP also exhibits substantially smaller accuracy drops on many other distribution shifts \citep{radford21a,jia2021scaling,basic2021}.

More formally, our experiments measure the accuracy of a model $f$ on two test distributions $D_1$ and $D_2$, which we abbreviate as $\text{acc}_{D_1}(f)$ and $\text{acc}_{D_2}(f)$.
Usually $D_1$ is the ImageNet (ILSVRC-2012) test set and $D_2$ is one of multiple out-of-distribution test sets.
An ideal model would achieve close to 100\% accuracy on both distributions.
Since such machine models currently do not exist, we instead have to compare the robustness of models with varying accuracies across the two distributions.
In these comparisons, an important confounder is that simply increasing accuracy on distribution $D_1$ often already results in accuracy gains on $D_2$ \citep{taori2020measuring,miller2021accuracy}.
For instance, Figure \ref{fig:figure1} shows a range of ImageNet models (blue points) in a scatter plot with ImageNet accuracy on the $x$-axis ($\text{acc}_{D_1}(f)$) and accuracy under distribution shift on the $y$-axis ($\text{acc}_{D_2}(f)$).
The models achieve higher accuracy under distribution shift just by virtue of having higher ImageNet accuracy.

In order to address the confounder of ImageNet accuracy when evaluating robustness, \citet{taori2020measuring} quantified robustness as \emph{accuracy beyond the baseline} given by ImageNet models.
The authors called this quantity \emph{effective robustness}.
In Figure \ref{fig:figure1}, effective robustness corresponds to the vertical lift of a model above the blue baseline given by ImageNet-trained models.
\citet{radford21a} then demonstrated that their CLIP models achieve high effective robustness (the purple line).
Mathematically, we first fit a baseline function $\beta: \mathbb{R} \rightarrow \mathbb{R}$ that maps from the accuracy $\text{acc}_{D_1}(f)$ of baseline models $f$ to the corresponding $\text{acc}_{D_2}(f)$.
For a new model $f'$, the effective robustness is then given by $\rho(f') = \text{acc}_{D_2}(f') - \beta(\text{acc}_{D_1}(f'))$.
This is the main quantity we visualize in this paper to understand the robustness of CLIP models.

Similar to \citet{taori2020measuring} and \citet{radford21a}, we focus on \emph{natural} distribution shifts, which arise from natural variations such as lighting, geographic location, crowdsourcing process, etc.
Natural distribution shifts stand in contrast to \emph{synthetic} distribution shifts, where an existing test set is intentionally computationally modified to reduce model accuracy (e.g., by adding Gaussian noise, blur, or adversarial perturbations).
Since natural distribution shifts resemble real data, we choose the following popular distribution shifts:\footnote{See Appendix \ref{app:distribution_examples} for examples of the shifts. In some figures we omit ImageNet-A due to the piecewise linear response created by the adversarial filtering process. We refer the reader to \citet{taori2020measuring} for details.}
\begin{enumerate}
\item ImageNet-V2 \citep{recht2019imagenet}: a reproduction of the ImageNet validation set with distribution shift due to changes in the crowdsourcing process.
\item ImageNet-Sketch \citep{wang2019learning}: black and white sketches of ImageNet images.
\item ImageNet-R \citep{hendrycks2021many}: renditions (e.g., art, patterns, etc.) of 200 ImageNet classes.
\item ObjectNet \citep{objectnet}: real-world objects from ImageNet with crowd-sourced random backgrounds, rotations, and viewpoints.
\item ImageNet-A \citep{hendrycks2019natural}: naturally occurring examples filtered so they are misclassified by a ResNet-50 model.
\end{enumerate}

An important property of effective robustness on these distribution shifts is that only varying the \emph{size} of the training set (holding its \emph{distribution} constant) does \emph{not} influence effective robustness.
In particular, \citet{taori2020measuring} and \citet{miller2021accuracy} showed that randomly sub-sampling the training set changes the accuracy, but not the effective robustness of the resulting models.
This rules out the training set size as a cause for CLIP's high effective robustness (the training set size is still important for the ImageNet \emph{accuracy} of the CLIP models).

\subsection{Additional related work}
Language image pre-training has been an active area of research for multiple years, including initial contributions such as VirTex \citep{Desai021}, ICMLM \citep{SariyildizPL20}, and ConVIRT \citep{ZhangConvirt}.
\citet{radford21a} and \citet{jia2021scaling} continued this line of work and trained on significantly larger datasets to achieve competitive performance on a variety of tasks, as well as obtain models with unprecedented robustness.

Related recent work also studies exactly \emph{where} the generalization capabilities of CLIP come from.
\citet{devillers2021does} investigate whether models that use multimodal information (such as text \& images) have superior generalization capabilities --
as measured by few-shot and linear probe performance -- to models that use only one type of information (images or text).
Their analysis found that for both few-shot and linear probe settings there was no consistent advantage of multimodal models over models using only a single modality.
In contrast, our work studies the \emph{robustness} of CLIP and how language specifically affects its capability to generalize out of distribution.
An important difference between our experiments and those of \citet{devillers2021does} is that we control for in-distribution accuracy in our comparison between the models to separate accuracy and robustness.
Furthermore, \citet{andreassen2021evolution} study the effect of fine-tuning on robustness.
They find that effective robustness decreases almost monotonically during the fine-tuning process, pointing to the zero-shot capability of CLIP as a source of its robustness.

Since the original CLIP paper \citep{radford21a}, there have been a series of follow up works, including
ALIGN \cite{jia2021scaling}, BASIC \cite{basic2021} and LiT \cite{lit}, each of which has made
contributions to improving either the robustness or base accuracy of large pre-trained image-text models.
Most related to our experiments in Section \ref{sec:yfcc} is LiT, which uses a pre-trained image
model and fine-tunes only the text head of the language-image model to achieve high accuracy on downstream tasks.
However, we note that this work differs from our contribution in that LiT still fine-tunes a language model on a dataset of 4 billion image-caption pairs
to achieve its zero-shot capability, while we simply convert the captions to class labels using substring matching and train a regular image classifier.

\paragraph{Image captioning datasets.}
Existing literature offers a variety of public datasets with image-text pairs.
Examples range from medium to large scale, including
MS-COCO \cite{chen2015microsoft},
SBU \cite{ordonez2011im2text},
Conceptual Captions 3M \cite{sharma2018conceptual} and 12M \cite{changpinyo2021conceptual},
RedCaps \cite{desai2021redcaps},
WIT \cite{srinivasan2021wit},
YFCC 100M \cite{thomee2016yfcc100m} and
LAION 400M \cite{schuhmann2021laion}.
Compared to these datasets, ImageNet-Captions contains high quality
classification labels along with text associated with each
image.
Moreover, ImageNet-Captions is designed such that the distribution
of images strongly resembles that of ImageNet, which is
widely used for training and evaluating models, enabling
controlled experiments such as comparisons between
multiclass supervised training and image-text training.

\section{\imagenetcaptions}
\begin{figure}[bt]
    \centering
    \includegraphics[width=0.7\textwidth]{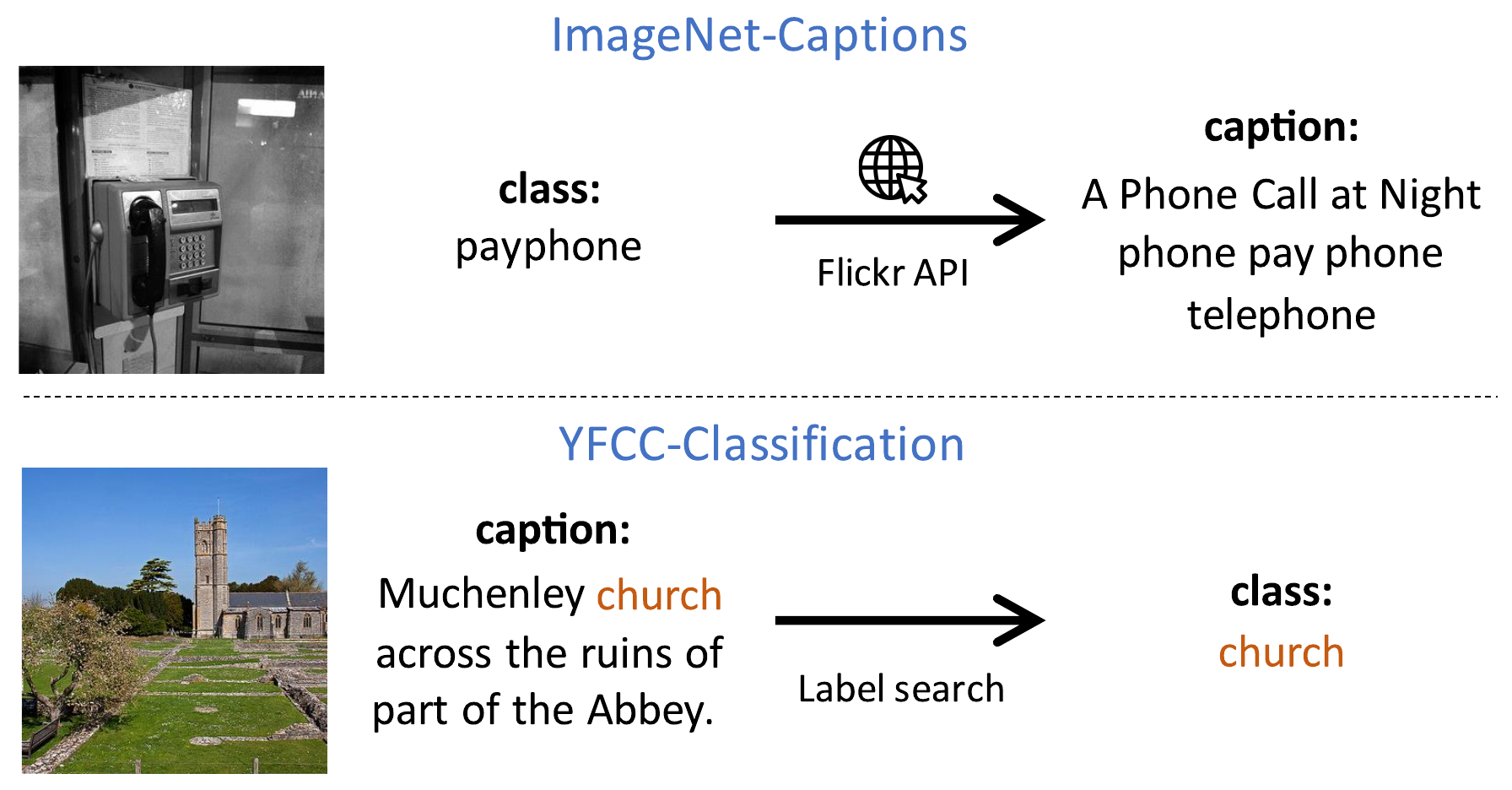}
    \caption{\label{fig:imagenet-yfcc}
    Overview of the two main training sets in our experiments.
    (Top) We introduce the ImageNet-Captions dataset, where we \emph{augment} a subset of the ImageNet 2012 training set images with the corresponding original captions collected from Flickr.
    (Bottom) We convert the YFCC image-caption dataset into YFCC-Classification by searching for class labels in the YFCC captions and then \emph{removing} the text annotations.
    These two datasets allows us to evaluate the impact of language-image training on robustness because we can compare language-image training with standard classification training \emph{on the same set of images}.}
\end{figure}
\begin{figure*}[h]
    \captionsetup[subfigure]{labelformat=empty}
    \centering
    \begin{subfigure}[t]{.30\linewidth}
        \centering
        \captionsetup{justification=centering}
        \includegraphics[width=0.7\linewidth]{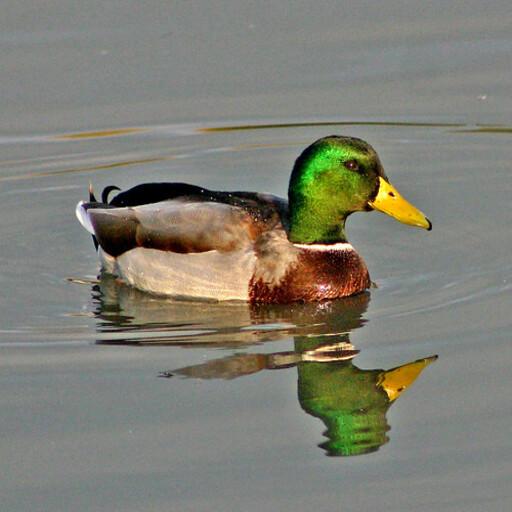}
        \caption{Title: \texttt{Reflected Duck} \\ Description: \\ Tags: \texttt{lake}, \texttt{water}, \texttt{bird} [6 tags omitted]}
    \end{subfigure}
    \begin{subfigure}[t]{0.30\linewidth}
        \centering
        \captionsetup{justification=centering}
        \includegraphics[width=0.7\linewidth]{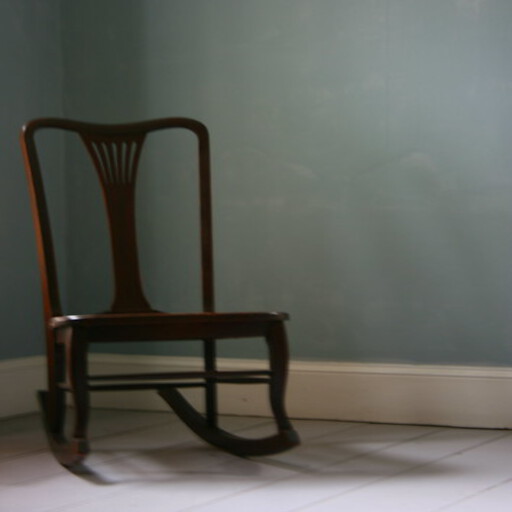}
        \caption{Title: \texttt{SILENT ROCKER} \\ Description: \texttt{MOSE'S MOTHER HAS LEFT THE BuILDING} [10 words omitted]\\ Tags: \texttt{rockingchair}, \texttt{rock}, \texttt{chair} [2 tags omitted]}
    \end{subfigure}
    \begin{subfigure}[t]{0.30\linewidth}
        \centering
        \captionsetup{justification=centering}
        \includegraphics[width=0.7\linewidth]{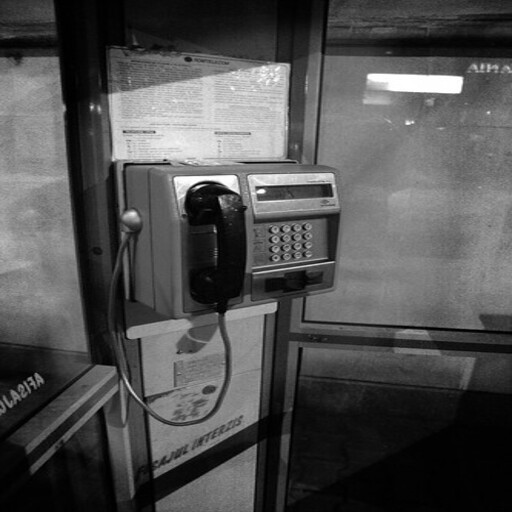}
        \caption{Title: \texttt{A Phone Call at Night} \\ Description: \texttt{I might have a thing with telephones} [174 words omitted] \\ Tags: \texttt{phone}, \texttt{telephone}, \texttt{blackandwhite} [7 tags omitted] }
    \end{subfigure}
    \caption{\label{fig:sample} Three sample images from ImageNet-Captions. Their respective ImageNet labels are:  drake, rocking chair, payphone.}
\end{figure*}
We now describe ImageNet-Captions\footnote{The dataset is available at \url{https://github.com/mlfoundations/imagenet-captions}.}, our new dataset for experiments with image-text supervision.
Four desiderata guided the creation of ImageNet-Captions:
\begin{enumerate}
\item To isolate the effect of natural language supervision on effective robustness, we require a dataset that contains \emph{both} natural language supervision and
traditional classification labels.
This setup allows us to train classifiers separately with contrastive image-text losses and with standard classification losses on the \emph{same images} and compare the resulting models.
Differences in the models are then solely due to different loss functions, not architectural differences or different training distributions.
\item The text annotations in the dataset should come from the original image source, as opposed to synthetically generated captions from curated templates or an image captioning model.
This helps ensure that the dataset is representative of image-text data ``in the wild'' and minimizes artifacts from templates or machine models.
\item The dataset should be related to commonly studied benchmarks, such as ImageNet, in order to have good baselines and comparable training methods.
\item The dataset should be large enough to support training on contemporary neural networks.
\end{enumerate}
Before our paper, no dataset satisfied these constraints.

We constructed ImageNet-Captions to satisfy all four desiderata.
ImageNet-Captions is a subset of the ImageNet Large Scale Visual Recognition Challenge (ILSVRC) 2012 training set, paired with the original image title, description, and tags from Flickr (recall that a large part of ImageNet was sourced from the Flickr image hosting website).
Figure \ref{fig:sample} shows three sample image-text pairs from our dataset.

\subsection{Constructing ImageNet-Captions}
Since ImageNet is a widely used image classification benchmark, our goal was to augment the 2012 ImageNet training set with original text data.
A priori, this is a difficult task since the standard 2012 ImageNet release does not contain any metadata for the images.
As a starting point, we leveraged three facts about ImageNet:
\begin{itemize}
\item A large fraction of ImageNet is sourced from Flickr.
\item The ImageNet fall 2011 release contained URLs for each image in the full ImageNet dataset.
\item For a given photo identifier, the Flickr API provides the associated text data.
\end{itemize}

Our dataset construction began with filtering the 14,197,122 image URLs in the ImageNet fall 2011 release to only include images from Flickr.
In addition, we restricted the images to just the 1,000 classes included in the 2012 ImageNet competition (every entry in the fall 2011 release contains both a URL and a class label).
After this filtering, we were left with 642,147 images belonging to 999 classes (all classes in ILSVRC-2012 except ``teddy bear'').

Next, we ran the image deduplication routine of \citet{idealods2019imagededup} to remove images that were not in the ILSVRC-2012 training set.
In addition, we removed text containing profanity.
This left us with a dataset of 463,622 images that are in the ILSVRC-2012 training set, along with the newly obtained corresponding text data.
In particular, for each image we extracted a title (the text at the top of the Flickr image), description (the text at the bottom of the Flickr image), and user-provided tags.
Since these images are a subset of ILSVRC-2012, we also have a corresponding class label that can be used for standard ImageNet training.

\subsection{Properties of ImageNet-Captions}
\begin{table}[t]
    \centering
    \rowcolors{2}{}{gray!25}
  \caption{\label{tab:imagenet_captions} Images from {\imagenetcaptions} contain three types of metadata: titles, description, and tags.
  For each type of metadata, this table shows the number of images that have corresponding metadata that contains the class label of the image.
  For most images, the class label is in at least one text field, indicating that \imagenetcaptions \ is suitable for language-image training.\\[-.1cm]}
  \centering
    \begin{tabular}{lcc}
    \toprule
    Caption Type               & \# Images & \% of Total \\\midrule
    Title Only                 & 239,495   & 51.6 \\
    Description Only           & 134,387   & 28.9 \\
    Tags Only                  & 342,340   & 73.8 \\
    Title, Tag and Description & 435,239   & 93.8 \\\bottomrule
    \end{tabular}
\end{table}

The resulting dataset contains captions from a mix of 127 different languages with the bulk (90\%) coming from English.
We further inspected the quality of image-text pairs by checking for the presence of the desired class label in the associated text.
Table \ref{tab:imagenet_captions} summarizes the analysis.
We find that for 94\% of the images, the name of the ImageNet class is present in the corresponding text.
This indicates that most of the captions contain relevant information about the class and are suitable for training image-text models.
For additional statistics, see Appendix \ref{app:imagenet_captions_statistics}.

\section{ImageNet-Captions experiments}
\label{sec:experiments}
In this section, we use the ImageNet-Captions dataset to investigate the effect of language on robustness. ImageNet-Captions provides a simple comparison with vision-only methods because ImageNet is considered the premier benchmark for image classification. We train the ResNet-50 based CLIP model on ImageNet-Captions with a contrastive loss, as well as the vision encoder of that CLIP model with an additional linear layer on the equivalent image classification dataset. Training details are in Appendix \ref{app:imagenet_captions_train}.

\begin{table*}[t]
\centering
\rowcolors{3}{}{gray!25}
\caption{Evaluating different caption variants across ImageNet (IN) natural distribution shifts.
Results are reported in top-1 accuracy (\%).
The best performing caption uses title, tags, and description.
Although the language filter makes captions cleaner, the decrease in overall dataset size decreases performance.}
\vskip 0.03in
\label{table:caption_variants}
\begin{tabular}{cccccccccccc}
\toprule
Title & Desc & Tags & Filter
& Size & Relative & IN & IN-V2 & IN-R  & IN Sketch & ObjectNet & IN-A  \\
&&&&& Size (\%) &&&&&&\\
\midrule
\cmark &        &        & \cmark & 197K & 42.6 & 15.7 & 12.2 & 6.6 & 1.1 & 5.5 & 2.4 \\
\cmark &        &        &        & 459K & 99.0 & 26.2 & 20.7 & 9.5 & 2.6 & 8.4 & 2.7 \\
\cmark & \cmark &        & \cmark & 312K & 67.4 & 21.9 & 16.5 & 8.0 & 1.7 & 6.1 & 2.2 \\
\cmark & \cmark &        &        & 461K & 99.4 & 27.8 & 21.6 & 9.6 & 3.0 & 8.0 & 2.7 \\
\cmark & \cmark & \cmark & \cmark & 367K & 79.3 & 26.5 & 20.3 & 8.9 & 2.3 & 7.9 & 2.5 \\
\cmark & \cmark & \cmark &        & 464K & 100.0 & 31.5 & 24.0 & 10.9 & 2.7 & 9.1 & 3.0 \\
\bottomrule
\end{tabular}
\vskip -0.1in
\end{table*}

\subsection{Caption construction}
When constructing ImageNet-Captions, we had to choose which parts of the metadata to include in the caption.
To do so, we ran experiments on variants that included just the title, the title followed by the description, and the title followed by the tags followed by the description.
Furthermore, \citet{radford21a} use a filter to keep only images with captions in English. We create additional variants of the dataset by applying a similar filter. As shown in Table \ref{table:caption_variants}, captions that include more information appear to perform better. Furthermore, it seems that filtering for cleaner captions does not make up for the loss of image-caption pairs.
In caption construction ablations, images with empty captions were dropped, causing variation in dataset size across the experiments in Table \ref{table:caption_variants}.

\subsection{Robustness}

\begin{figure*}[bt]
    \centering
    \includegraphics[width=\textwidth]{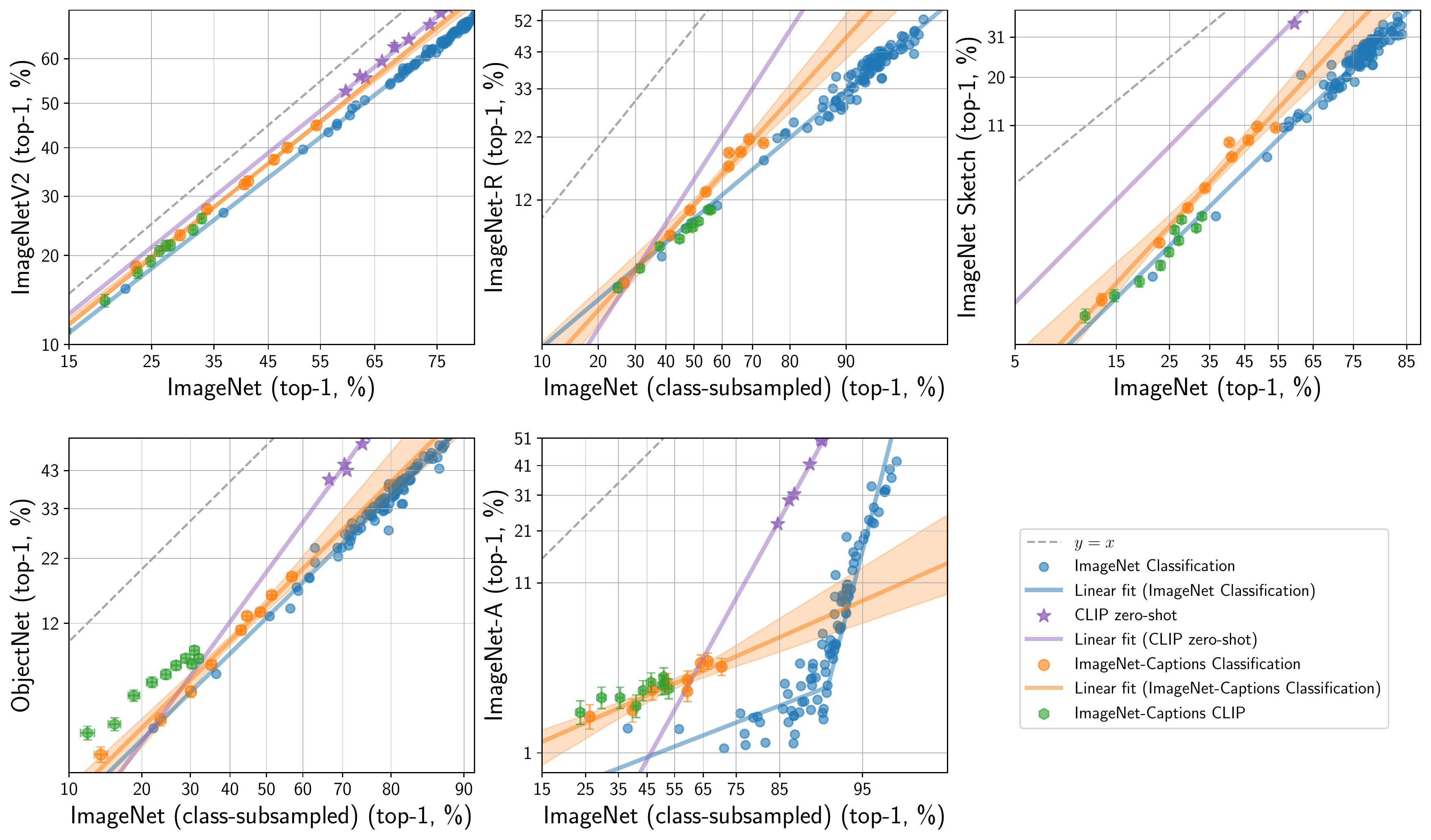}
    \caption{\label{fig:imagenet_captions_base} On most natural distribution shifts, models trained with language information from ImageNet-Captions follow the same trend as models trained without it. Neither comes close to achieving the robustness of OpenAI's CLIP models. }
\end{figure*}

To determine the robustness of models trained on ImageNet-Captions, we evaluate on ImageNet and compare with natural distribution shifts in ImageNetV2, ImageNet-R, ImageNet Sketch, ObjectNet, and ImageNet-A.
In Figure \ref{fig:imagenet_captions_base}, we see that ImageNet-Captions CLIP models roughly follow the same linear trends as ImageNet-Captions classification models across the various distribution shifts. This shows that CLIP models are not more robust than classification models trained on the same dataset, despite the difference of language supervision. This is a better comparison than that with ImageNet classification models because there is no longer the potential confounding factor of the datasets having different image distributions. Nevertheless, these models do not achieve the robustness seen in CLIP models from \citet{radford21a}. Additional experiment details can be found in Appendix \ref{app:imagenet_captions_subsampled} and \ref{app:imagenet_captions_classification}.

\subsection{Pre-training on language}
\begin{figure}[h]
    \centering
    \includegraphics[width=0.48\textwidth]{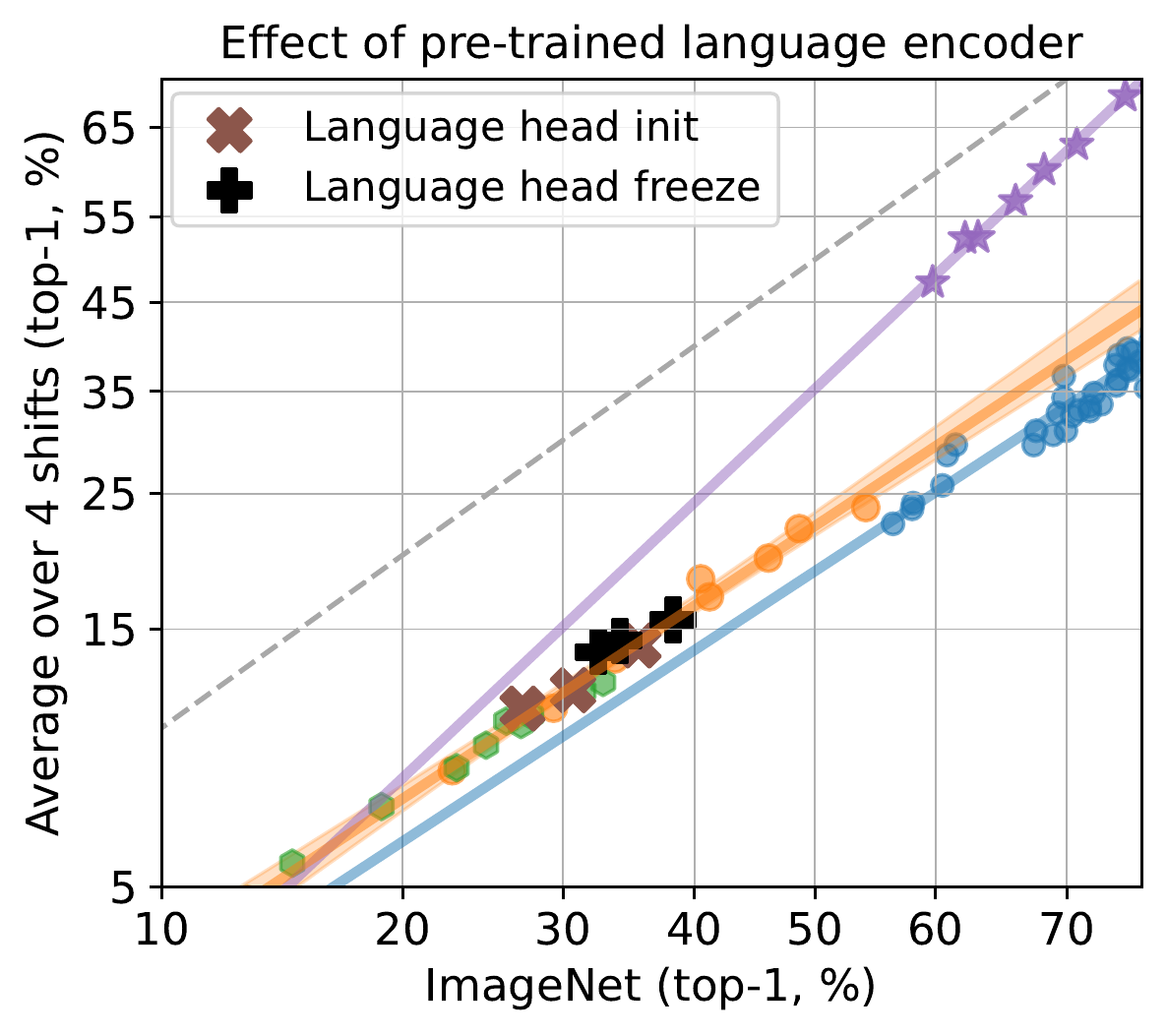}
    \caption{\label{fig:language_variation} Using the weights from OpenAI’s pre-trained CLIP model does not improve robustness, despite the large size of the full CLIP training set (400 million images). This is further evidence that language supervision does not increase robustness. See Figure \ref{fig:imagenet_captions_base} for remaining legend elements.}
\end{figure}

While the above experiments show that language supervision from ImageNet-Captions does not contribute to a model's robustness, it does not rule out robustness coming from the language supervision of OpenAI's proprietary dataset used to train CLIP. Therefore we ran additional experiments where we loaded the pre-trained OpenAI CLIP model onto the language encoder, while randomly initializing the vision encoder. We trained ImageNet-Captions on this setup, with an additional variant where we also freeze the language encoder's weights. As seen in Figure \ref{fig:language_variation}, while both the unfrozen and frozen variants of the pre-trained language encoder increased the accuracy of the model when compared to the completely randomly initialized model, neither variant provided additional effective robustness. Detailed experiment results can be found in Appendix \ref{app:imagenet_captions_language}.

\subsection{Effect of using templates}
Given that images in ImageNet-Captions have a corresponding ImageNet class, we can try to leverage this information to investigate the effect of captions and class information on both accuracy and robustness. \citet{radford21a} introduces prompt templates in formats similar to ``A photo of a \{label\}." Creating templates for ImageNet-Captions is different than doing so for other image-text datasets because each image already has an assigned label; for other datasets, creating a template requires looking through the caption for classes, which are not guaranteed to be in the caption.

We found that attaching templates at the beginning of captions (followed by Title+Tags+Description) achieves 34.7\% ImageNet top-1 accuracy, which is 3.2\% more than without the templates. However, using the templates by themselves as the captions achieves 50.5\% ImageNet top-1 accuracy, suggesting that the additional information in the captions hurts ImageNet performance. The model trained on the equivalent classification task achieves 48.7\%, which suggests that with additional parameter tuning, classification may be similar to CLIP training on templates.
Detailed experiment results can be found in Appendix \ref{app:imagenet_captions_templates}.

While using templates instead of captions can increase ImageNet performance, it does not improve robustness. Figure~\ref{fig:template_robustness} in Appendix~\ref{app:imagenet_captions_templates} shows that using templates on top of the captions follows linear trends similar to ImageNet-Captions. In fact, training a model on all of ImageNet using templates behaves like an equivalent classification model.

\subsection{Improving ImageNet performance using captions}
While language supervision does not improve robustness, it is still possible that the additional information may improve ImageNet accuracy. We investigate this by running experiments on ImageNet, augmented with ImageNet-Captions. It is well known that ResNet-50 achieves 77.15\% top-1 ImageNet accuracy \cite{he2016deep}. As a similar baseline, we achieve 76.62\% top-1 ImageNet accuracy by training the CLIP model with templates as the caption.

We have tried improving this baseline by initializing the language head with the OpenAI pre-trained model, using the combined ImageNet (templates) and ImageNet-Captions for training, using ImageNet-Captions as text augmentation when available, and contrasting the image encoding with both the template and the ImageNet-Captions caption encodings when available. However, all of these fall within $\pm$1\% of the baseline. Note that concatenating the captions to the templates changes the image distributions, while some of the other approaches do not. On the other hand, restricting the images used to those within ImageNet-Captions hints that language may help improve ImageNet performance.

Appendix \ref{app:imagenet_captions_variation} presents detailed results.
The experiments we have run are non-exhaustive, and we leave it to future work to find whether language information can improve ImageNet performance, and more broadly, vision task performance.

\section{YFCC experiments}
\label{sec:yfcc}
Our experiments in the previous section show that language supervision alone does not improve robustness.
To further understand the source of CLIP's robustness, we now investigate whether it is possible to train a representation with minimal or even no language supervision that still yields the same robustness as CLIP.
These results will provide further evidence that CLIP's robustness stems from the more diverse data distribution, not the presence of language supervision.

Our experiments in this section start with a language-image training set on which CLIP exhibits improved robustness: the Yahoo Flickr Creative Commons dataset (YFCC)~\cite{thomee2016yfcc100m}.
To test whether the image data in YFCC alone can improve robustness, we contrastively pre-train a ``standard'' image representation on YFCC that does not involve the language part of the dataset.
Building on this image-only representation, we then train a zero-shot classifier with only minimal text processing (substring matches).
The resulting classifier achieves effective robustness close to CLIP.
This demonstrates that the training distribution, not language supervision at training time, is the main reason behind CLIP’s robustness.

\textbf{Dataset.}
In this section, we use the YFCC-15M~\cite{radford21a} dataset, a subset of YFCC-100M~\cite{thomee2016yfcc100m} filtered to only images with English titles or descriptions.
The dataset contains 14,829,396 images with natural language captions associated with each image.

To train image classifiers on YFCC-15M, we convert YFCC-15M into a classification dataset with class labels for each image, which we denote \yfcccls.
We assign ImageNet labels to each image using a simple strategy: if the title or description contains the name of an ImageNet synset or synonym \citep{wordnet}, we assign the corresponding synset label to the image.
If an image contains no or multiple ImageNet synsets, we discard that image.
This results in 1,694,125 images (11.4\% of the full dataset) covering 953 ILSVRC classes. The least common class has 1 image, while the most common has 280,351 images.

\textbf{Classification training.}
We use a ViT-Base (ViT-B/16) model fine-tuned using the softmax cross-entropy loss on \yfcccls.
Since this data is only a fraction of YFCC-15M, we initialize the classification model with a SimCLR model pre-trained on YFCC-15M from \citet{mu2021slip}.
\Cref{app:noclip} includes implementation details and ablations.

\begin{table}[t]
    \centering
    \vspace{-.2cm}
    \caption{\label{tab:clip_yfcc} Comparing CLIP training with (language model free) classification models on YFCC-15M. All experiments use a ViT-B/16 backbone. The CLIP results are from \citet{mu2021slip}.
    Image-only contrastive learning followed by a simple text matching stage for classification nearly matches the performance of CLIP with a full language model.}
    \begin{tabular}{lcc}
        \toprule
        Training style & ImageNet & Avg OOD \\
        \midrule
        CLIP & 37.9 & 19.9 \\
        SimCLR $\rightarrow$ Classification & 35.7 & 18.8 \\
        \bottomrule
    \end{tabular}
\end{table}

\textbf{Results.}
We present our results in \Cref{tab:clip_yfcc} and \Cref{fig:figure1}.
A CLIP model trained on all images and captions from YFCC-15M yields an ImageNet top-1 accuracy of 37.9\%.
Our baseline classification model\footnote{We call this baseline ``NoCLIP'', for ``\textbf{No}w we use SimCLR+Classification instead of \textbf{C}ontrastive \textbf{L}anguage-\textbf{I}mage \textbf{P}re-training}, which trains SimCLR on YFCC-15M, but fine-tunes on only a small fraction (about 11\%) of the supervision in YFCC-15M, results in an accuracy of 35.7\%, which we found surprisingly close to CLIP.
Further, as shown in \Cref{fig:figure1} (``YFCC SimCLR + Classification''), our baseline model's effective robustness is similar to that of CLIP.

Appendix \ref{app:yfcc_multi} provides figures that plot the above results on various distribution shifts, as well as a model trained on \yfcccls\ from scratch.
Since the training set is now about nine times smaller than YFCC-15M, the resulting models trained from scratch achieve much lower accuracy and are hard to compare to CLIP.

Overall, we find that despite largely eschewing language, and training on a fraction of the supervision, our baseline model results in high effective robustness, similar to CLIP.
These results indicate that image-only pre-training followed by classification fine-tuning can match the robustness of CLIP, and that language pre-training is \emph{not} necessary for effective robustness.
Models trained on YFCC consistently achieve higher effective robustness than models trained on ImageNet, which shows that different training distributions have different levels of effective robustness.

\section{Effect of test time prompts}
\label{sec:testprompts}
\vspace{-.1cm}
\begin{figure*}[h]
\begin{center}
\centerline{\includegraphics[width=\textwidth]{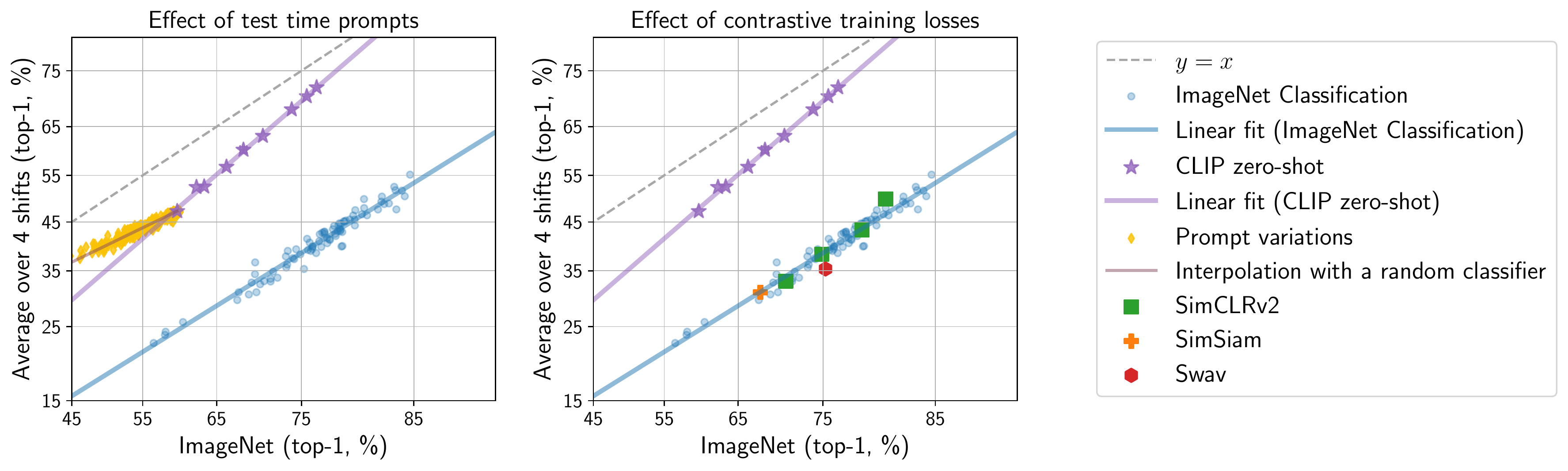}}
\caption{ Effect of prompting strategies and contrastive objectives on
robustness.
(Left) On most natural distribution shifts, effect of
prompting on effective robustness is similar to that of
random interpolation. (Right) Models pre-trained with various
contrastive objectives on ImageNet do not achieve the same effective robustness as
CLIP models.}
\label{fig:prompt_selfsup}
\end{center}
\end{figure*}

As another hypothesis, we study whether natural language prompts affect CLIP's robustness.
Recall that prompts consist of a template (e.g., \textit{``a photo of \_\_\_''}) and the name of a class in the dataset.
\citet{radford21a} showed how to use multiple templates by averaging their text representations.
Similarly, it is also possible to use multiple class names for each class if synonyms exist (e.g. \textit{microwave} and \textit{microwave oven}).
To investigate the influence of specific prompts in the robustness of CLIP, we conduct a series of experiments using a trained CLIP model and multiple prompting strategies.
Specifically, we vary:

\begin{itemize}
    \item The templates used, using one of the following three options:

        \quad i) Templates from \citet{radford21a};

        \quad ii) No templates (i.e., only the class names);

        \quad iii) Random words appended before and after the class name.\footnote{Templates are composed by one to ten random words along with the class name, in an arbitrary position. Random words are drawn using \url{https://pypi.org/project/Random-Word/}.}
    \item The names of the classes, using one of the following three sources:

        \quad i) Class names from \citet{radford21a};

        \quad ii) Class names from WordNet synset \cite{wordnet};

        \quad iii) A combination of the previous two sources.

    \item The number of templates used, chosen from $\{1, 2, 4, 8, 16, 32, 80\}$.
    \item The maximum synonyms per class, one of $\{1, 2, 4\}$.
\end{itemize}

Figure \ref{fig:prompt_selfsup} (left) shows the results from over a hundred experiments.
We find that specific choices of prompts can have a substantial impact on performance. %
While some prompt variations did increase effective robustness, this increase is entirely due to the substantially reduced accuracy.
In particular, one can achieve the same change in effective robustness by simply interpolating with a random classifier (which sees no performance change under distribution shift).
We illustrate this behavior with the brown line in Figure \ref{fig:prompt_selfsup} (left).
Overall, these results show that prompts are not the source of robustness of CLIP models.

\section{Effect of contrastive training losses}
\label{sec:loss}

Finally, we explore contrastive pre-training as a potential source of CLIP's robustness.
Contrastive pre-training is a popular method for self-supervised representation
learning that encourages similar pairs to be close and dissimilar pairs to be far apart in a learned representation space.
In \citet{radford21a}, the similar pairs are images and their
corresponding captions.
In SimCLR \cite{simclr2020}, similar pairs are
the same images with different data augmentation.

As the contrastive loss is core to CLIP's approach,
we explore whether
contrastive approaches independently promote effective robustness.
Figure~\ref{fig:prompt_selfsup} (right) shows results for various popular contrastive
methods, including SimCLRv2 \cite{chen2020big}, SimSiam \cite{chen2020exploring} and SwAV \cite{caron2020unsupervised},
pre-trained on ImageNet.
We evaluate on ImageNet and the five distribution shifts.
While the methods differ significantly from each other (e.g.,
different augmentation strategies,
memory banks, and feature clustering techniques),
they consistently exhibit little to no effective robustness.
Figure~\ref{fig:prompt_selfsup} leaves out ImageNet-A because of its piecewise behavior.
See Appendix \ref{app:selfsup} for further details.

\section{Conclusion}
\label{sec:conclusion}
The previous sections have systematically ruled out the training set size, language supervision, and the contrastive loss function as explanations for the large robustness gains achieved by the CLIP models of \citet{radford21a}.
In addition, Section \ref{sec:yfcc} has demonstrated that changing the training distribution from ImageNet(-Captions) to YFCC substantially affects the robustness of the resulting models.
We arrive at a clear conclusion: CLIP's robustness is dominated by the choice of the training distribution, with other factors playing a small or non-existent role.
While language supervision is still helpful for easily assembling training sets, it is not the primary driver for robustness.

Our paper connects the fields of robustness, learning from language \& vision, and data-centric machine learning.
Moreover, our results add to a growing body of evidence that the training distribution plays a central role for mitigating real-world distribution shifts.
Hence, we believe that the sometimes overlooked area of dataset design offers a promising avenue for increasing the robustness of machine learning models, and we hope that the community invests more efforts into this direction.

\section{Acknowledgements}
We would like to thank
Wieland Brendel,
Nicholas Carlini,
Yair Carmon,
Rahim Entezari,
Tatsunori Hashimoto,
Jong Wook Kim,
Hongseok Namkoong,
Alec Radford,
and Rohan Taori
for valuable conversations while working on this project.
This work is in part supported by the NSF AI Institute for Foundations of Machine Learning (IFML) and Open Philanthropy.

{\small
\bibliographystyle{plainnat}
\bibliography{references}

\begin{thebibliography}{39}
\providecommand{\natexlab}[1]{#1}
\providecommand{\url}[1]{\texttt{#1}}
\expandafter\ifx\csname urlstyle\endcsname\relax
  \providecommand{\doi}[1]{doi: #1}\else
  \providecommand{\doi}{doi: \begingroup \urlstyle{rm}\Url}\fi

\bibitem[Andreassen et~al.(2021)Andreassen, Bahri, Neyshabur, and
  Roelofs]{andreassen2021evolution}
Anders Andreassen, Yasaman Bahri, Behnam Neyshabur, and Rebecca Roelofs.
\newblock The evolution of out-of-distribution robustness throughout
  fine-tuning.
\newblock 2021.
\newblock \url{https://arxiv.org/abs/2106.15831}.

\bibitem[Barbu et~al.(2019)Barbu, Mayo, Alverio, Luo, Wang, Gutfreund,
  Tenenbaum, and Katz]{objectnet}
Andrei Barbu, David Mayo, Julian Alverio, William Luo, Christopher Wang, Dan
  Gutfreund, Josh Tenenbaum, and Boris Katz.
\newblock Objectnet: A large-scale bias-controlled dataset for pushing the
  limits of object recognition models.
\newblock In \emph{Advances in Neural Information Processing Systems
  (NeurIPS)}, 2019.
\newblock
  \url{https://proceedings.neurips.cc/paper/2019/file/97af07a14cacba681feacf3012730892-Paper.pdf}.

\bibitem[Caron et~al.(2020)Caron, Misra, Mairal, Goyal, Bojanowski, and
  Joulin]{caron2020unsupervised}
Mathilde Caron, Ishan Misra, Julien Mairal, Priya Goyal, Piotr Bojanowski, and
  Armand Joulin.
\newblock Unsupervised learning of visual features by contrasting cluster
  assignments.
\newblock 2020.
\newblock \url{https://arxiv.org/abs/2006.09882}.

\bibitem[Changpinyo et~al.(2021)Changpinyo, Sharma, Ding, and
  Soricut]{changpinyo2021conceptual}
Soravit Changpinyo, Piyush Sharma, Nan Ding, and Radu Soricut.
\newblock Conceptual 12m: Pushing web-scale image-text pre-training to
  recognize long-tail visual concepts.
\newblock In \emph{Proceedings of the IEEE/CVF Conference on Computer Vision
  and Pattern Recognition}, 2021.
\newblock \url{https://arxiv.org/abs/2102.08981}.

\bibitem[Chen et~al.(2020{\natexlab{a}})Chen, Kornblith, Norouzi, and
  Hinton]{simclr2020}
Ting Chen, Simon Kornblith, Mohammad Norouzi, and Geoffrey~E. Hinton.
\newblock A simple framework for contrastive learning of visual
  representations.
\newblock In \emph{Proceedings of the 37th International Conference on Machine
  Learning, {ICML} 2020, 13-18 July 2020, Virtual Event}, volume 119 of
  \emph{Proceedings of Machine Learning Research}. {PMLR}, 2020{\natexlab{a}}.
\newblock \url{http://proceedings.mlr.press/v119/chen20j.html}.

\bibitem[Chen et~al.(2020{\natexlab{b}})Chen, Kornblith, Swersky, Norouzi, and
  Hinton]{chen2020big}
Ting Chen, Simon Kornblith, Kevin Swersky, Mohammad Norouzi, and Geoffrey
  Hinton.
\newblock Big self-supervised models are strong semi-supervised learners.
\newblock 2020{\natexlab{b}}.
\newblock \url{https://arxiv.org/abs/2006.10029}.

\bibitem[Chen and He(2020)]{chen2020exploring}
Xinlei Chen and Kaiming He.
\newblock Exploring simple siamese representation learning. corr abs/2011.10566
  (2020).
\newblock 2020.
\newblock \url{https://arxiv.org/abs/2011.10566}.

\bibitem[Chen et~al.(2015)Chen, Fang, Lin, Vedantam, Gupta, Doll{\'a}r, and
  Zitnick]{chen2015microsoft}
Xinlei Chen, Hao Fang, Tsung-Yi Lin, Ramakrishna Vedantam, Saurabh Gupta, Piotr
  Doll{\'a}r, and C~Lawrence Zitnick.
\newblock Microsoft coco captions: Data collection and evaluation server.
\newblock 2015.
\newblock \url{https://arxiv.org/abs/1504.00325}.

\bibitem[Cubuk et~al.(2020)Cubuk, Zoph, Shlens, and Le]{cubuk2020randaugment}
Ekin~D Cubuk, Barret Zoph, Jonathon Shlens, and Quoc~V Le.
\newblock Randaugment: Practical automated data augmentation with a reduced
  search space.
\newblock In \emph{Conference on Computer Vision and Pattern Recognition
  Workshops}, 2020.
\newblock \url{https://arxiv.org/abs/1909.13719}.

\bibitem[Desai and Johnson(2021)]{Desai021}
Karan Desai and Justin Johnson.
\newblock Virtex: Learning visual representations from textual annotations.
\newblock In \emph{{IEEE} Conference on Computer Vision and Pattern
  Recognition, {CVPR} 2021, virtual, June 19-25, 2021}, pages 11162--11173.
  Computer Vision Foundation / {IEEE}, 2021.
\newblock URL
  \url{https://openaccess.thecvf.com/content/CVPR2021/html/Desai\_VirTex\_Learning\_Visual\_Representations\_From\_Textual\_Annotations\_CVPR\_2021\_paper.html}.

\bibitem[Desai et~al.(2021)Desai, Kaul, Aysola, and Johnson]{desai2021redcaps}
Karan Desai, Gaurav Kaul, Zubin~Trivadi Aysola, and Justin Johnson.
\newblock Redcaps: Web-curated image-text data created by the people, for the
  people.
\newblock 2021.
\newblock \url{https://arxiv.org/abs/2111.11431}.

\bibitem[Devillers et~al.(2021)Devillers, Choksi, Bielawski, and
  VanRullen]{devillers2021does}
Benjamin Devillers, Bhavin Choksi, Romain Bielawski, and Rufin VanRullen.
\newblock Does language help generalization in vision models?
\newblock 2021.
\newblock \url{https://arxiv.org/abs/2104.08313}.

\bibitem[Djolonga et~al.(2021)Djolonga, Yung, Tschannen, Romijnders, Beyer,
  Kolesnikov, Puigcerver, Minderer, D'Amour, Moldovan, Gelly, Houlsby, Zhai,
  and Lucic]{Djolonga_2021_CVPR}
Josip Djolonga, Jessica Yung, Michael Tschannen, Rob Romijnders, Lucas Beyer,
  Alexander Kolesnikov, Joan Puigcerver, Matthias Minderer, Alexander D'Amour,
  Dan Moldovan, Sylvain Gelly, Neil Houlsby, Xiaohua Zhai, and Mario Lucic.
\newblock On robustness and transferability of convolutional neural networks.
\newblock In \emph{Proceedings of the IEEE/CVF Conference on Computer Vision
  and Pattern Recognition (CVPR)}, June 2021.
\newblock \url{https://arxiv.org/abs/2007.08558}.

\bibitem[Dosovitskiy et~al.(2020)Dosovitskiy, Beyer, Kolesnikov, Weissenborn,
  Zhai, Unterthiner, Dehghani, Minderer, Heigold, Gelly, Uszkoreit, and
  Houlsby]{vit2020}
Alexey Dosovitskiy, Lucas Beyer, Alexander Kolesnikov, Dirk Weissenborn,
  Xiaohua Zhai, Thomas Unterthiner, Mostafa Dehghani, Matthias Minderer, Georg
  Heigold, Sylvain Gelly, Jakob Uszkoreit, and Neil Houlsby.
\newblock An image is worth 16x16 words: Transformers for image recognition at
  scale.
\newblock \emph{CoRR}, abs/2010.11929, 2020.
\newblock \url{https://arxiv.org/abs/2010.11929}.

\bibitem[He et~al.(2016)He, Zhang, Ren, and Sun]{he2016deep}
Kaiming He, Xiangyu Zhang, Shaoqing Ren, and Jian Sun.
\newblock Deep residual learning for image recognition.
\newblock In \emph{2016 {IEEE} Conference on Computer Vision and Pattern
  Recognition, {CVPR} 2016, Las Vegas, NV, USA, June 27-30, 2016}, pages
  770--778. {IEEE} Computer Society, 2016.
\newblock \doi{10.1109/CVPR.2016.90}.
\newblock \url{https://doi.org/10.1109/CVPR.2016.90}.

\bibitem[He et~al.(2021)He, Chen, Xie, Li, Doll{\'a}r, and
  Girshick]{he2021masked}
Kaiming He, Xinlei Chen, Saining Xie, Yanghao Li, Piotr Doll{\'a}r, and Ross
  Girshick.
\newblock Masked autoencoders are scalable vision learners.
\newblock 2021.
\newblock \url{https://arxiv.org/abs/2111.06377}.

\bibitem[Heckel and Yilmaz(2021)]{heckel2021early}
Reinhard Heckel and Fatih~Furkan Yilmaz.
\newblock Early stopping in deep networks: Double descent and how to eliminate
  it.
\newblock In \emph{International Conference on Learning Representations}, 2021.
\newblock URL \url{https://openreview.net/forum?id=tlV90jvZbw}.

\bibitem[Hendrycks et~al.(2019)Hendrycks, Zhao, Basart, Steinhardt, and
  Song]{hendrycks2019natural}
Dan Hendrycks, Kevin Zhao, Steven Basart, Jacob Steinhardt, and Dawn Song.
\newblock Natural adversarial examples.(2019).
\newblock 2019.
\newblock \url{https://arxiv.org/abs/1907.07174}.

\bibitem[Hendrycks et~al.(2021)Hendrycks, Basart, Mu, Kadavath, Wang, Dorundo,
  Desai, Zhu, Parajuli, Guo, et~al.]{hendrycks2021many}
Dan Hendrycks, Steven Basart, Norman Mu, Saurav Kadavath, Frank Wang, Evan
  Dorundo, Rahul Desai, Tyler Zhu, Samyak Parajuli, Mike Guo, et~al.
\newblock The many faces of robustness: A critical analysis of
  out-of-distribution generalization.
\newblock In \emph{Proceedings of the IEEE/CVF International Conference on
  Computer Vision}, 2021.
\newblock \url{https://arxiv.org/abs/2006.16241}.

\bibitem[Jain et~al.(2019)Jain, Lennan, John, and Tran]{idealods2019imagededup}
Tanuj Jain, Christopher Lennan, Zubin John, and Dat Tran.
\newblock Imagededup.
\newblock \url{https://github.com/idealo/imagededup}, 2019.

\bibitem[Jia et~al.(2021)Jia, Yang, Xia, Chen, Parekh, Pham, Le, Sung, Li, and
  Duerig]{jia2021scaling}
Chao Jia, Yinfei Yang, Ye~Xia, Yi-Ting Chen, Zarana Parekh, Hieu Pham, Quoc~V.
  Le, Yunhsuan Sung, Zhen Li, and Tom Duerig.
\newblock Scaling up visual and vision-language representation learning with
  noisy text supervision, 2021.
\newblock \url{https://arxiv.org/abs/2102.05918}.

\bibitem[Miller(1995)]{wordnet}
George~A. Miller.
\newblock Wordnet: A lexical database for english.
\newblock \emph{Commun. ACM}, Nov 1995.
\newblock \url{https://doi.org/10.1145/219717.219748}.

\bibitem[Miller et~al.(2021)Miller, Taori, Raghunathan, Sagawa, Koh, Shankar,
  Liang, Carmon, and Schmidt]{miller2021accuracy}
John~P Miller, Rohan Taori, Aditi Raghunathan, Shiori Sagawa, Pang~Wei Koh,
  Vaishaal Shankar, Percy Liang, Yair Carmon, and Ludwig Schmidt.
\newblock Accuracy on the line: on the strong correlation between
  out-of-distribution and in-distribution generalization.
\newblock In \emph{International Conference on Machine Learning}. PMLR, 2021.
\newblock \url{https://arxiv.org/abs/2107.04649}.

\bibitem[Mu et~al.(2021)Mu, Kirillov, Wagner, and Xie]{mu2021slip}
Norman Mu, Alexander Kirillov, David Wagner, and Saining Xie.
\newblock Slip: Self-supervision meets language-image pre-training.
\newblock 2021.
\newblock \url{https://arxiv.org/abs/2112.12750}.

\bibitem[Ordonez et~al.(2011)Ordonez, Kulkarni, and Berg]{ordonez2011im2text}
Vicente Ordonez, Girish Kulkarni, and Tamara Berg.
\newblock Im2text: Describing images using 1 million captioned photographs.
\newblock \emph{Advances in neural information processing systems}, 24, 2011.
\newblock
  \url{https://proceedings.neurips.cc/paper/2011/file/5dd9db5e033da9c6fb5ba83c7a7ebea9-Paper.pdf}.

\bibitem[Pham et~al.(2021)Pham, Dai, Ghiasi, Liu, Yu, Luong, Tan, and
  Le]{basic2021}
Hieu Pham, Zihang Dai, Golnaz Ghiasi, Hanxiao Liu, Adams~Wei Yu, Minh{-}Thang
  Luong, Mingxing Tan, and Quoc~V. Le.
\newblock Combined scaling for zero-shot transfer learning.
\newblock \emph{CoRR}, abs/2111.10050, 2021.
\newblock \url{https://arxiv.org/abs/2111.10050}.

\bibitem[Radford et~al.(2021)Radford, Kim, Hallacy, Ramesh, Goh, Agarwal,
  Sastry, Askell, Mishkin, Clark, Krueger, and Sutskever]{radford21a}
Alec Radford, Jong~Wook Kim, Chris Hallacy, Aditya Ramesh, Gabriel Goh,
  Sandhini Agarwal, Girish Sastry, Amanda Askell, Pamela Mishkin, Jack Clark,
  Gretchen Krueger, and Ilya Sutskever.
\newblock Learning transferable visual models from natural language
  supervision.
\newblock In Marina Meila and Tong Zhang, editors, \emph{Proceedings of the
  38th International Conference on Machine Learning, {ICML} 2021, 18-24 July
  2021, Virtual Event}, volume 139 of \emph{Proceedings of Machine Learning
  Research}. {PMLR}, 2021.
\newblock \url{http://proceedings.mlr.press/v139/radford21a.html}.

\bibitem[Recht et~al.(2019)Recht, Roelofs, Schmidt, and
  Shankar]{recht2019imagenet}
Benjamin Recht, Rebecca Roelofs, Ludwig Schmidt, and Vaishaal Shankar.
\newblock Do imagenet classifiers generalize to imagenet?
\newblock In \emph{International Conference on Machine Learning}. PMLR, 2019.
\newblock \url{https://arxiv.org/abs/1902.10811}.

\bibitem[Russakovsky et~al.(2015)Russakovsky, Deng, Su, Krause, Satheesh, Ma,
  Huang, Karpathy, Khosla, Bernstein, et~al.]{russakovsky2015imagenet}
Olga Russakovsky, Jia Deng, Hao Su, Jonathan Krause, Sanjeev Satheesh, Sean Ma,
  Zhiheng Huang, Andrej Karpathy, Aditya Khosla, Michael Bernstein, et~al.
\newblock Imagenet large scale visual recognition challenge.
\newblock \emph{International journal of computer vision}, 115\penalty0
  (3):\penalty0 211--252, 2015.

\bibitem[Sariyildiz et~al.(2020)Sariyildiz, Perez, and Larlus]{SariyildizPL20}
Mert~B{\"{u}}lent Sariyildiz, Julien Perez, and Diane Larlus.
\newblock Learning visual representations with caption annotations.
\newblock In Andrea Vedaldi, Horst Bischof, Thomas Brox, and Jan{-}Michael
  Frahm, editors, \emph{Computer Vision - {ECCV} 2020 - 16th European
  Conference, Glasgow, UK, August 23-28, 2020, Proceedings, Part {VIII}},
  volume 12353 of \emph{Lecture Notes in Computer Science}, pages 153--170.
  Springer, 2020.
\newblock \doi{10.1007/978-3-030-58598-3\_10}.
\newblock URL \url{https://doi.org/10.1007/978-3-030-58598-3\_10}.

\bibitem[Schuhmann et~al.(2021)Schuhmann, Vencu, Beaumont, Kaczmarczyk, Mullis,
  Katta, Coombes, Jitsev, and Komatsuzaki]{schuhmann2021laion}
Christoph Schuhmann, Richard Vencu, Romain Beaumont, Robert Kaczmarczyk,
  Clayton Mullis, Aarush Katta, Theo Coombes, Jenia Jitsev, and Aran
  Komatsuzaki.
\newblock Laion-400m: Open dataset of clip-filtered 400 million image-text
  pairs.
\newblock 2021.
\newblock \url{https://arxiv.org/abs/2111.02114}.

\bibitem[Shankar et~al.(2020)Shankar, Roelofs, Mania, Fang, Recht, and
  Schmidt]{shankar2020evaluating}
Vaishaal Shankar, Rebecca Roelofs, Horia Mania, Alex Fang, Benjamin Recht, and
  Ludwig Schmidt.
\newblock Evaluating machine accuracy on imagenet.
\newblock In \emph{International Conference on Machine Learning}. PMLR, 2020.
\newblock \url{https://proceedings.mlr.press/v119/shankar20c.html}.

\bibitem[Sharma et~al.(2018)Sharma, Ding, Goodman, and
  Soricut]{sharma2018conceptual}
Piyush Sharma, Nan Ding, Sebastian Goodman, and Radu Soricut.
\newblock Conceptual captions: A cleaned, hypernymed, image alt-text dataset
  for automatic image captioning.
\newblock In \emph{Proceedings of the 56th Annual Meeting of the Association
  for Computational Linguistics (Volume 1: Long Papers)}, 2018.

\bibitem[Srinivasan et~al.(2021)Srinivasan, Raman, Chen, Bendersky, and
  Najork]{srinivasan2021wit}
Krishna Srinivasan, Karthik Raman, Jiecao Chen, Michael Bendersky, and Marc
  Najork.
\newblock Wit: Wikipedia-based image text dataset for multimodal multilingual
  machine learning.
\newblock 2021.
\newblock \url{https://arxiv.org/abs/2103.01913}.

\bibitem[Taori et~al.(2020)Taori, Dave, Shankar, Carlini, Recht, and
  Schmidt]{taori2020measuring}
Rohan Taori, Achal Dave, Vaishaal Shankar, Nicholas Carlini, Benjamin Recht,
  and Ludwig Schmidt.
\newblock Measuring robustness to natural distribution shifts in image
  classification.
\newblock 2020.
\newblock \url{https://arxiv.org/abs/2007.00644}.

\bibitem[Thomee et~al.(2016)Thomee, Shamma, Friedland, Elizalde, Ni, Poland,
  Borth, and Li]{thomee2016yfcc100m}
Bart Thomee, David~A Shamma, Gerald Friedland, Benjamin Elizalde, Karl Ni,
  Douglas Poland, Damian Borth, and Li-Jia Li.
\newblock Yfcc100m: The new data in multimedia research.
\newblock \emph{Communications of the ACM}, 59\penalty0 (2):\penalty0 64--73,
  2016.

\bibitem[Wang et~al.(2019)Wang, Ge, Xing, and Lipton]{wang2019learning}
Haohan Wang, Songwei Ge, Eric~P Xing, and Zachary~C Lipton.
\newblock Learning robust global representations by penalizing local predictive
  power.
\newblock 2019.
\newblock \url{https://arxiv.org/abs/1905.13549}.

\bibitem[Zhai et~al.(2021)Zhai, Wang, Mustafa, Steiner, Keysers, Kolesnikov,
  and Beyer]{lit}
Xiaohua Zhai, Xiao Wang, Basil Mustafa, Andreas Steiner, Daniel Keysers,
  Alexander Kolesnikov, and Lucas Beyer.
\newblock Lit: Zero-shot transfer with locked-image text tuning.
\newblock \emph{CoRR}, 2021.
\newblock \url{https://arxiv.org/abs/2111.07991}.

\bibitem[Zhang et~al.(2020)Zhang, Jiang, Miura, Manning, and
  Langlotz]{ZhangConvirt}
Yuhao Zhang, Hang Jiang, Yasuhide Miura, Christopher~D. Manning, and Curtis~P.
  Langlotz.
\newblock Contrastive learning of medical visual representations from paired
  images and text.
\newblock \emph{CoRR}, abs/2010.00747, 2020.
\newblock URL \url{https://arxiv.org/abs/2010.00747}.

\end{thebibliography}


\begin{thebibliography}{2}
\providecommand{\natexlab}[1]{#1}
\providecommand{\url}[1]{\texttt{#1}}
\expandafter\ifx\csname urlstyle\endcsname\relax
  \providecommand{\doi}[1]{doi: #1}\else
  \providecommand{\doi}{doi: \begingroup \urlstyle{rm}\Url}\fi

\bibitem[He et~al.(2016)He, Zhang, Ren, and Sun]{he2016deep}
He, K., Zhang, X., Ren, S., and Sun, J.
\newblock Deep residual learning for image recognition.
\newblock In \emph{2016 {IEEE} Conference on Computer Vision and Pattern
  Recognition, {CVPR} 2016, Las Vegas, NV, USA, June 27-30, 2016}, pp.\
  770--778. {IEEE} Computer Society, 2016.
\newblock \doi{10.1109/CVPR.2016.90}.
\newblock URL \url{https://doi.org/10.1109/CVPR.2016.90}.

\bibitem[Radford et~al.(2021)Radford, Kim, Hallacy, Ramesh, Goh, Agarwal,
  Sastry, Askell, Mishkin, Clark, Krueger, and Sutskever]{radford21a}
Radford, A., Kim, J.~W., Hallacy, C., Ramesh, A., Goh, G., Agarwal, S., Sastry,
  G., Askell, A., Mishkin, P., Clark, J., Krueger, G., and Sutskever, I.
\newblock Learning transferable visual models from natural language
  supervision.
\newblock In Meila, M. and Zhang, T. (eds.), \emph{Proceedings of the 38th
  International Conference on Machine Learning, {ICML} 2021, 18-24 July 2021,
  Virtual Event}, volume 139 of \emph{Proceedings of Machine Learning
  Research}, pp.\  8748--8763. {PMLR}, 2021.
\newblock URL \url{http://proceedings.mlr.press/v139/radford21a.html}.

\end{thebibliography}
}

\newpage
\appendix
\addcontentsline{toc}{section}{Appendices}
\section{Distribution shift examples}
\label{app:distribution_examples}
\begin{figure*}[h]
    \centering
    \includegraphics[width=\textwidth]{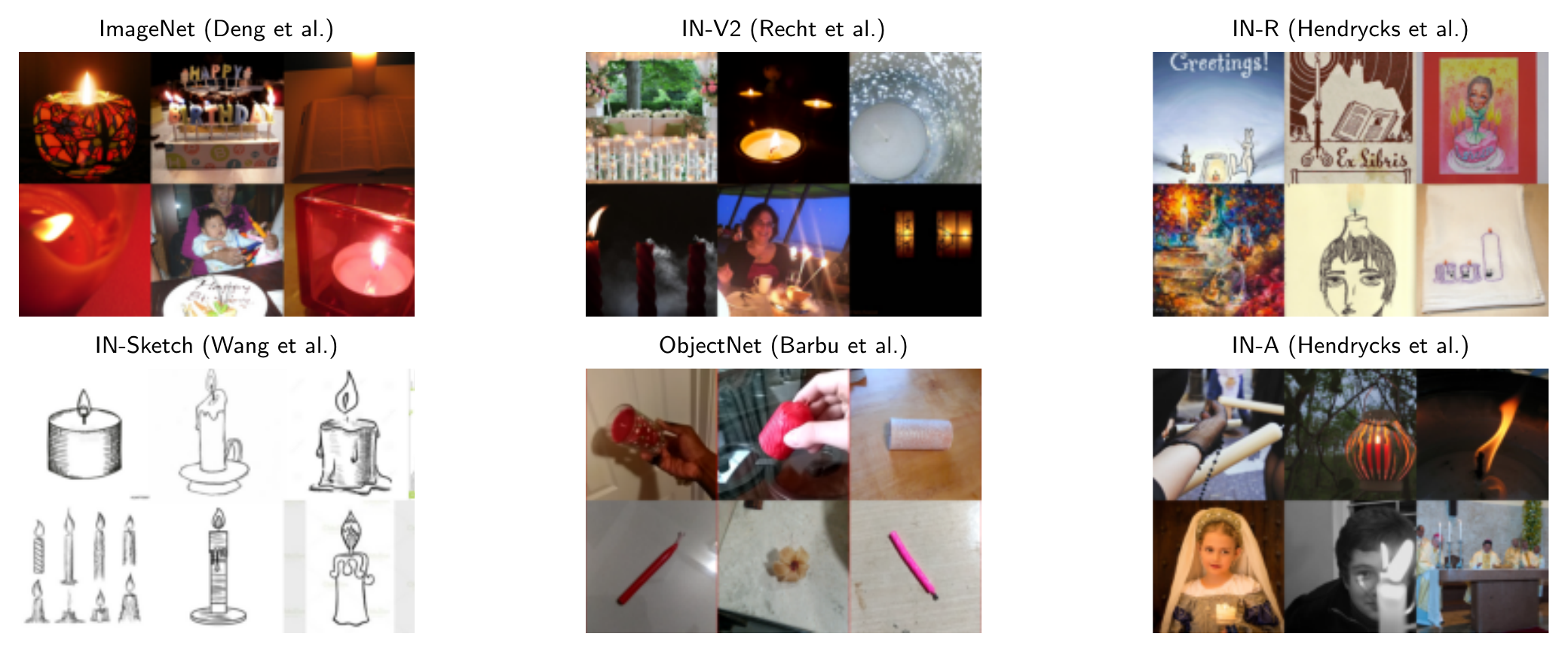}
    \caption{\label{fig:distribution_overview} Samples of the class candle from the various distribution shifts that we evaluate on in our experiments.}
\end{figure*}

\section{ImageNet-Captions experiments training details}
\label{app:imagenet_captions_train}

CLIP experiments are trained with cross-entropy losses using AdamW optimizer with initial learning rate of 0.001 and a cosine-annealing learning rate schedule with 500 warmup steps. Hyperparameters for AdamW are set at $\beta_1=0.9$, $\beta_2=0.999$, and $\epsilon=$1e-8. The batch size is set to 1024.
CLIP models trained on ImageNet-Captions are trained for 32 epochs, while ClIP models trained on all of ImageNet are trained for 90 epochs.

ImageNet-Captions classification models are trained with cross-entropy loss for 90 epochs using SGD with Nesterov momentum, setting weight decay to 0.0001, momentum to 0.9, and batch size to 256.
The initial learning rate is 0.1, and is decayed by 0.1 at epochs 30, 50, and 70.

The default augmentation is random resized crop to size 224 with scale set to $(0,9, 1.0)$, and then normalization. Additional augmentation indicates using random resized crop to size 224, random horizontal flips, and then normalization. Normalization is done with mean set to $(0.48145466, 0.4578275, 0.40821073)$, and standard deviation set to $(0.26862954, 0.26130258, 0.27577711)$.

\section{ImageNet-Captions CLIP subsampled experiments}
All models used here are the ResNet-50 based CLIP model used in \citet{radford21a}.
Experiments are trained on a class-balanced subset of ImageNet-Captions (IN-Captions).
\label{app:imagenet_captions_subsampled}
\begin{table*}[hbt!]
\centering
\rowcolors{4}{gray!25}{}
\begin{tabular}{lllllll}
\toprule
\cmidrule(r){1-2}
Experiment  & IN & IN-V2 & IN-R  & IN Sketch & ObjectNet & IN-A  \\
& (top-1, \%) & (top-1, \%) & (top-1, \%) & (top-1, \%) & (top-1, \%) & (top-1, \%) \\
\midrule
IN-Captions 30\% & 10.8 & 8.6 & 4.7 & 0.8 & 3.8 & 1.8 \\
IN-Captions 40\% & 14.7 & 11.2 & 5.8 & 1.0 & 4.2 & 2.2 \\
IN-Captions 50\% & 18.9 & 14.2 & 7.4 & 1.2 & 5.7 & 2.2 \\
IN-Captions 60\% & 23.0 & 17.6 & 8.0 & 1.6 & 6.6 & 2.0 \\
IN-Captions 70\% & 24.9 & 19.2 & 8.9 & 1.9 & 7.1 & 2.5 \\
IN-Captions 80\% & 27.1 & 21.4 & 9.0 & 2.2 & 7.8 & 2.8 \\
IN-Captions 100\% & 31.5 & 24.0 & 10.9 & 2.7 & 9.1 & 3.0 \\
\bottomrule
\end{tabular}
\vskip -0.1in
\end{table*}
\newpage

\section{ImageNet-Captions classification experiments}
Unless otherwise specified, all models used here are the ResNet-50 based visual encoder used in \citet{radford21a}, with an additional linear layer at the end.
Experiments are trained on a class-balanced subset of ImageNet-Captions (IN-Captions).
\label{app:imagenet_captions_classification}
\begin{table*}[hbt!]
\centering
\rowcolors{4}{gray!25}{}
\begin{tabular}{lllllll}
\toprule
\cmidrule(r){1-2}
Experiment  & IN & IN-V2 & IN-R  & IN Sketch & ObjectNet & IN-A  \\
& (top-1, \%) & (top-1, \%) & (top-1, \%) & (top-1, \%) & (top-1, \%) & (top-1, \%) \\
\midrule
IN-Captions 10\% & 12.9 & 10.0 & 5.0 & 1.0 & 3.0 & 1.7 \\
IN-Captions 20\% & 22.8 & 18.4 & 8.3 & 2.2 & 4.4 & 1.9 \\
IN-Captions 30\% & 29.3 & 23.1 & 10.8 & 3.6 & 6.0 & 2.5 \\
IN-Captions 40\% & 33.8 & 27.6 & 13.0 & 4.7 & 7.9 & 2.8 \\
IN-Captions 60\% & 41.2 & 33.0 & 16.7 & 7.2 & 11.2 & 2.9 \\
IN-Captions 80\% & 46.1 & 37.4 & 19.2 & 9.1 & 13.4 & 3.7 \\
IN-Captions 100\% & 48.7 & 40.0 & 21.6 & 10.8 & 15.8 & 3.8 \\
IN-Captions 100\%, Aug & 54.3 & 45.0 & 20.8 & 10.7 & 18.7 & 3.5 \\
ResNet-18, IN-Captions 100\% & 40.5 & 32.3 & 19.1 & 8.8 & 12.9 & 2.4 \\
\bottomrule
\end{tabular}
\vskip -0.1in
\end{table*}

\section{ImageNet-Captions language encoder experiments}
\label{app:imagenet_captions_language}
\begin{table*}[hbt!]
\centering
\rowcolors{5}{gray!25}{}
\begin{tabular}{llllllllll}
\toprule
Title & Desc & Tags & Filter & IN & IN-V2 & IN-R  & IN Sketch & ObjectNet & IN-A  \\
& & & & (top-1, \%) & (top-1, \%) & (top-1, \%) & (top-1, \%) & (top-1, \%) & (top-1, \%) \\
\midrule
\multicolumn{4}{l}{Language Initialized} &&&&&& \\
\midrule
\cmark &        &        & \cmark  & 19.9 & 15.3 & 8.4 & 1.9 & 6.5 & 2.2 \\
\cmark &        &        &         & 27.2 & 21.7 & 10.8 & 2.8 & 8.1 & 3.0 \\
\cmark & \cmark &        & \cmark  & 26.5 & 20.7 & 10.4 & 2.8 & 8.3 & 2.9 \\
\cmark & \cmark &        &         & 30.7 & 23.5 & 11.2 & 3.3 & 8.8 & 3.0 \\
\cmark & \cmark & \cmark & \cmark  & 31.2 & 24.0 & 12.0 & 3.2 & 10.0 & 2.8 \\
\cmark & \cmark & \cmark &         & 35.6 & 28.0 & 13.3 & 4.0 & 10.9 & 3.1 \\
\midrule
\rowcolor{white}
\multicolumn{4}{l}{Language Initialized and Frozen} &&&&&& \\
\midrule
\cmark &        &        & \cmark  & 23.4 & 18.5 & 11.0 & 3.2 & 8.6 & 2.9 \\
\cmark &        &        &         & 32.6 & 26.0 & 14.1 & 4.3 & 10.4 & 3.0 \\
\cmark & \cmark &        & \cmark  & 29.3 & 23.5 & 12.3 & 3.6 & 9.2 & 2.9 \\
\cmark & \cmark &        &         & 34.1 & 27.0 & 14.7 & 4.3 & 11.3 & 2.9 \\
\cmark & \cmark & \cmark & \cmark  & 35.4 & 27.5 & 14.7 & 4.8 & 11.0 & 3.3 \\
\cmark & \cmark & \cmark &         & 38.3 & 30.2 & 15.5 & 5.0 & 11.6 & 3.1 \\
\bottomrule
\end{tabular}
\vskip -0.1in
\end{table*}

\newpage

\section{ImageNet-Captions template experiments}
\label{app:imagenet_captions_templates}

\begin{table*}[hbt!]
\centering
\rowcolors{5}{}{gray!25}
\begin{tabular}{lllllll}
\toprule
\cmidrule(r){1-2}
Experiment & IN & IN-V2 & IN-R  & IN Sketch & ObjectNet & IN-A  \\
& (top-1, \%) & (top-1, \%) & (top-1, \%) & (top-1, \%) & (top-1, \%) & (top-1, \%) \\
\midrule
Title+Tags+Description (Base)          & 31.5 & 24.0 & 10.9 & 2.7 & 9.1 & 3.0 \\
Templates+Base       & 34.7 & 27.1 & 10.5 & 3.0 & 9.8 & 2.9 \\
Templates+Base, Aug  & 42.1 & 32.9 & 11.2 & 3.6 & 10.8 & 2.5 \\
Base as Templates                      & 50.5 & 39.6 & 17.4 & 7.5 & 13.9 & 3.4 \\
Base as Templates, Aug                 & 59.0 & 47.6 & 18.3 & 8.4 & 16.1 & 3.2 \\
Base as Classification                 & 48.7 & 40.2 & 21.5 & 10.8 & 15.7 & 3.7 \\
Base as Classification, Aug            & 54.2 & 45.0 & 20.7 & 10.6 & 18.7 & 3.6 \\
\bottomrule
\end{tabular}
\vskip -0.1in
\end{table*}

\begin{figure*}[h]
    \centering
    \includegraphics[width=\textwidth]{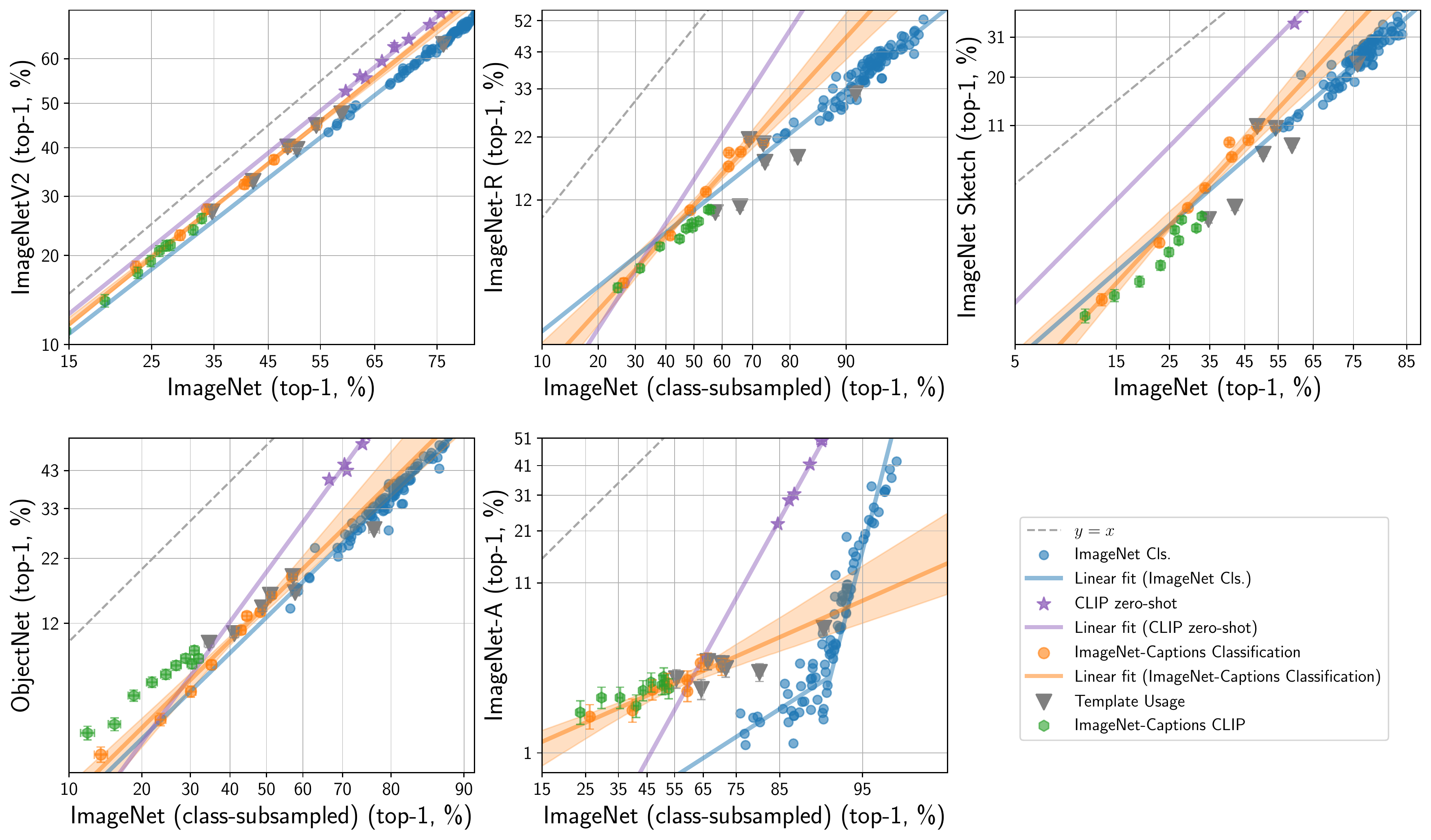}
    \caption{\label{fig:template_robustness} For the experiments that use templates in Appendix \ref{app:imagenet_captions_templates}, the templates do not increase a model's robustness.}
\end{figure*}

\newpage

\section{ImageNet templates and captions training variation experiments}
\label{app:imagenet_captions_variation}
Experiments in this section use all of ImageNet with the class labels replaced with templates.
Experiments that mention captions use captions for the subset of ImageNet in ImageNet-Captions.
\begin{table*}[hbt!]
\centering
\rowcolors{3}{}{gray!25}
\begin{tabular}{ll}
\toprule
\cmidrule(r){1-2}
Experiment  & ImageNet (top-1, \%) \\
\midrule
ImageNet using Templates (Base)   & 76.6 \\
Base, Language Initialized  & 76.6 \\
Base concatenated with Captions & 76.8 \\
Base concatenated with Captions, Language Initialized & 76.9 \\
Base, Captions as Text Augmentation & 76.4 \\
Base, Contrast with Template and Captions & 76.0 \\

\bottomrule
\end{tabular}
\vskip -0.1in
\end{table*}

\section{Self-supervised training variation experiments}
\label{app:selfsup}

\begin{figure}[h]
\vskip 0.2in
\begin{center}
\centerline{\includegraphics[width=0.9\columnwidth]{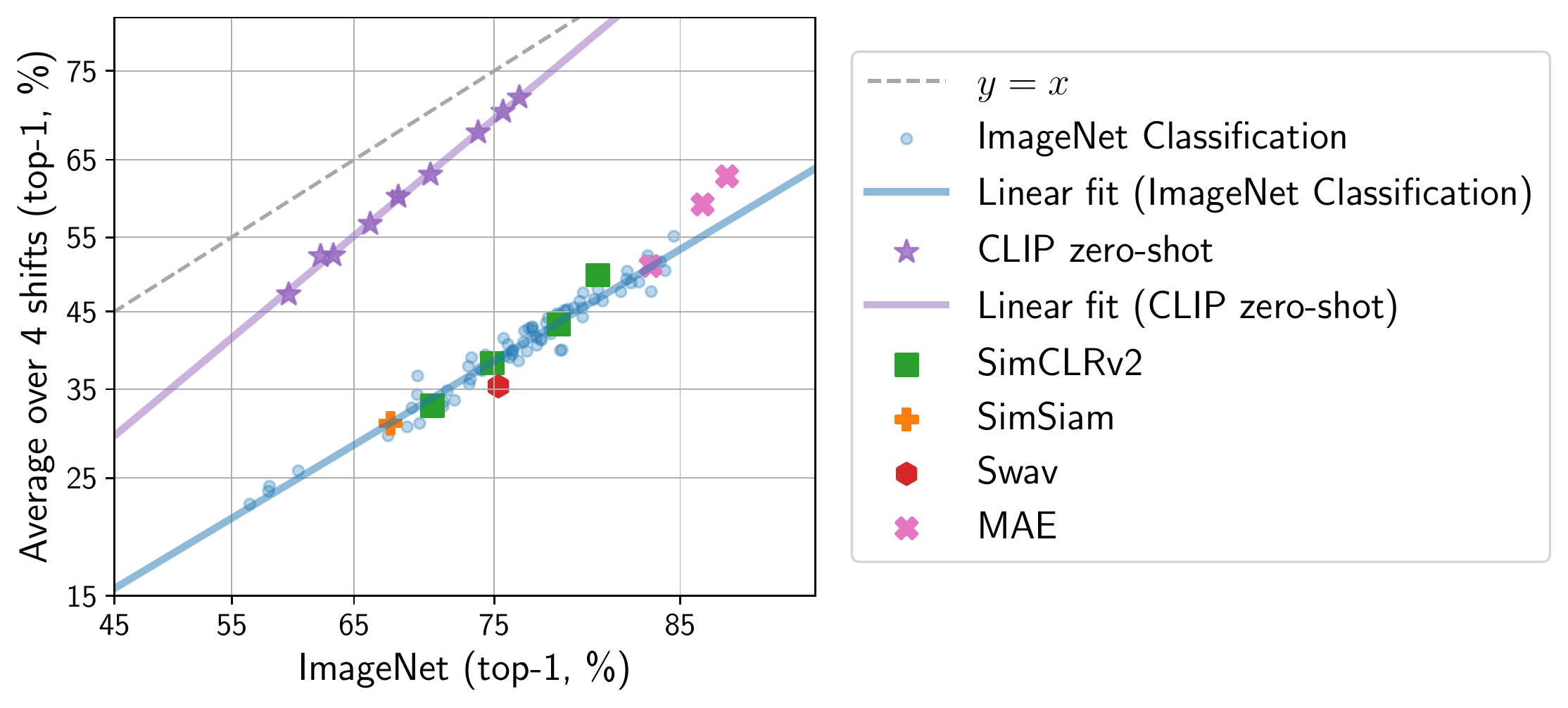}}
\caption{On most natural distribution shifts, models pre-trained on ImageNet with various contrastive objectives do not achieve effective robustness. y-axis is averaged over ImageNetV2, ImageNet-R, ImageNet Sketch, and ObjectNet. ImageNet-A is left out due to its piecewise function behavior.}
\label{fig:selfsup_mae}
\end{center}
\vskip -0.2in
\end{figure}

Figure~\ref{fig:selfsup_mae} supplements our analysis
of self-supervised methods with the recent MAE models of~\cite{he2021masked}.
In contrast to SimCLRV2 or MoCo, MAE is not a contrastive
training approach. While MAE provides more effective
robustness than other approaches, it is still much less
robust than CLIP.
However, the source of this robustness is an open question.

\begin{figure}[h]
\vskip 0.2in
\begin{center}
\centerline{\includegraphics[width=0.9\columnwidth]{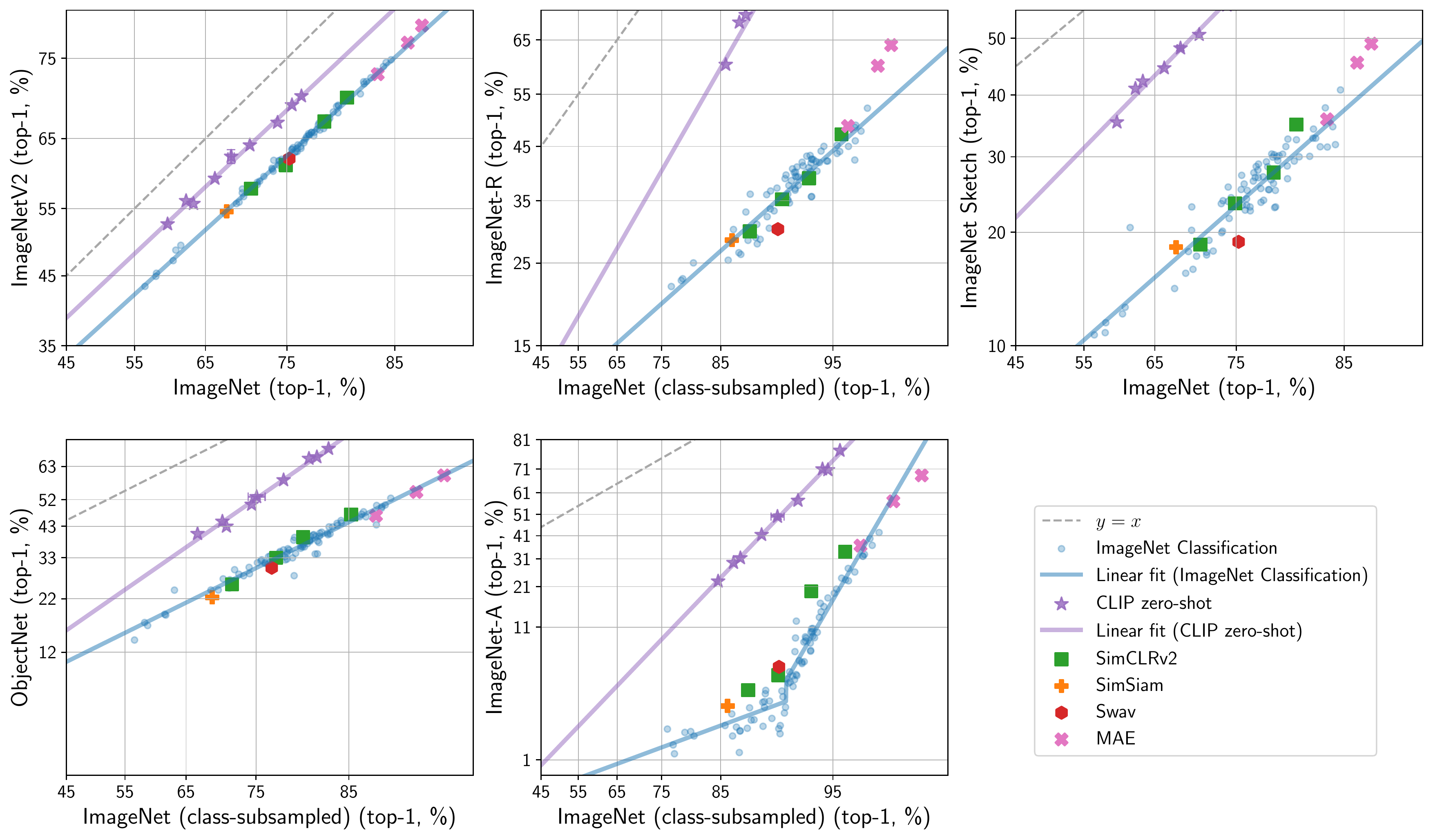}}
\caption{We repeat the prior figure except we plot each distribution shift separately, as well as include ImageNet-A.}
\label{fig:selfsup_mae_multi}
\end{center}
\vskip -0.2in
\end{figure}

\newpage

\section{Data cleaning}
\label{app:profanity}

The dataset is cleaned by removing samples thought to be offensive or profane.
More specifically, the captions are inspected via three independent mechanisms:
i) matches using a list of bad words and expressions;
ii) a profanity detector model trained on human-labeled samples;
iii) and human annotations, when applicable.

For the first filtering step, we use \small{\texttt{better-profanity}} library.\footnote{\url{https://pypi.org/project/better-profanity/}}
The list of words and expressions is initialized from the 835 default expressions from the library.\footnote{\url{https://github.com/snguyenthanh/better_profanity/blob/master/better_profanity/profanity_wordlist.txt}}
The authors manually reviewed these 835 expressions, finding 18 that could potentially be associated with ImageNet classes in non-profane captions.
Captions containing any of these 18 expressions were marked for subsequent human validation,
while captions that contained any of the remaining 817 expressions were automatically excluded.
This step is responsible for the largest portion of filtered samples, around 14 thousand samples (approximately 3\% of the data).
We now list the 18 expressions (\textbf{warning, the following words might be offensive}): \textit{breasts, cock, cocks, coon, cowgirl, dyke, nappy, nipple, nipples, organ, paddy, pot, sandbar, screw, screwed, screwing, sniper, titi}.

Data is additionally filtered using the \small{\texttt{profanity-check}} library.\footnote{\url{https://pypi.org/project/profanity-check/}}
The library detects profane or offensive language using a linear SVM model trained on 200 thousand human-labeled samples.
We use a threshold of 0.95, which filters 482 samples.

Finally, remaining captions that are found to contain any of the 18 expressions listed above are manually reviewed.
A total of 114 samples were found to be offensive or profane, and were removed from the dataset.

Combined, the three filtering steps filter 14,322 samples, approximately 3\% of the data.

\section{\baseline{} ablations}
\label{app:noclip}

We train NoCLIP using RandAug~\cite{cubuk2020randaugment} augmentation with N=3, magnitude=9,and magnitude std=0.5.
We use a cosine annealing learning rate initialized at \texttt{1e-3} with no warmup, and a batch size of 64, with a class-balanced sampler, training for 1 epoch. We use early stopping because training for more epochs hurts performance. \citet{heckel2021early}
showed that early stopping can be helpful when there is label noise.
We present ablations with different hyperparameters in \Cref{tab:noclip_ablations}.

\begin{table*}[h]
    \centering
    \caption{\baseline{} ablations.
    ``Label match'' indicates whether we search for the ImageNet synset or the synset and synonyms in the YFCC captions.
    ``Init'' indicates whether we train from scratch, or using SimCLR pre-training.
    ``Augmentation'' indicates different augmentation strategies; for ``RandAug'' we use N=3, magnitude=9, and magnitude std=0.5.
    ``Sampler'' is either class-balanced (`Class-bal') or `Random'(i.i.d., no balancing).
    }
    \rowcolors{3}{gray!25}{}
    \begin{tabular}{cccccc}
        \toprule
        Label match & Init & Augmentation & Sampler & Epochs & ImageNet (top-1, \%) \\\midrule
        Synset  & Scratch & Crop, Flip, Jitter & Class-bal         & 1 &  3.6 \\
        Synset  & Scratch & Crop, Flip, Jitter & Class-bal         & 20 &  5.7 \\\midrule
        Synset  & SimCLR & Crop, Flip, Jitter & Random         & 1 & 15.0 \\\midrule
        Synset  & SimCLR & Crop, Flip, Jitter & Class-bal. & 1 & 32.1 \\
        Synset  & SimCLR & Crop, Flip, Jitter & Class-bal. & 2 & 30.4 \\
        Synset  & SimCLR & Crop, Flip, Jitter & Class-bal. & 5 & 27.1 \\\midrule
        Synset + Synonyms & SimCLR & Crop, Flip, Jitter & Class-bal. & 1 & 34.5 \\
        Synset + Synonyms & SimCLR & RandAug              & Class-bal. & 1 & 35.7 \\
        \bottomrule
    \end{tabular}
    \label{tab:noclip_ablations}
\end{table*}

\section{YFCC-15M-Cls classes}
There were 47 ImageNet classes that did not show up in the YFCC captions.
They are the following:

tiger shark, boa constrictor, partridge, bee eater, crane bird, sea lion, toy terrier, Black and Tan Coonhound, English foxhound, Otterhound, Curly-coated Retriever, Brittany dog, Kuvasz, Groenendael dog, Greater Swiss Mountain Dog, Entlebucher Sennenhund, brussels griffon, tiger cat, tiger beetle, guinea pig, bath towel, bell tower, cassette player, cliff dwelling, construction crane, espresso machine, fountain pen, French horn, harp, one-piece bathing suit, measuring cup, missile, oxygen mask, plate rack, radio telescope, rain barrel, balaclava ski mask, slide rule, steel drum, totem pole, waffle iron, whiskey jug, window screen, Windsor tie, acorn squash, bell pepper, gyromitra
\newpage

\section{YFCC-15M-Cls additional experiments}
In addition to the experiments that trained on top of a model pre-trained on YFCC-15M, we also train a model on YFCC-15M-Cls from scratch. Note that this model is at a much lower accuracy regime than the rest of the models we look at.
\label{app:yfcc_multi}

\begin{figure*}[h]
    \centering
    \includegraphics[width=\textwidth]{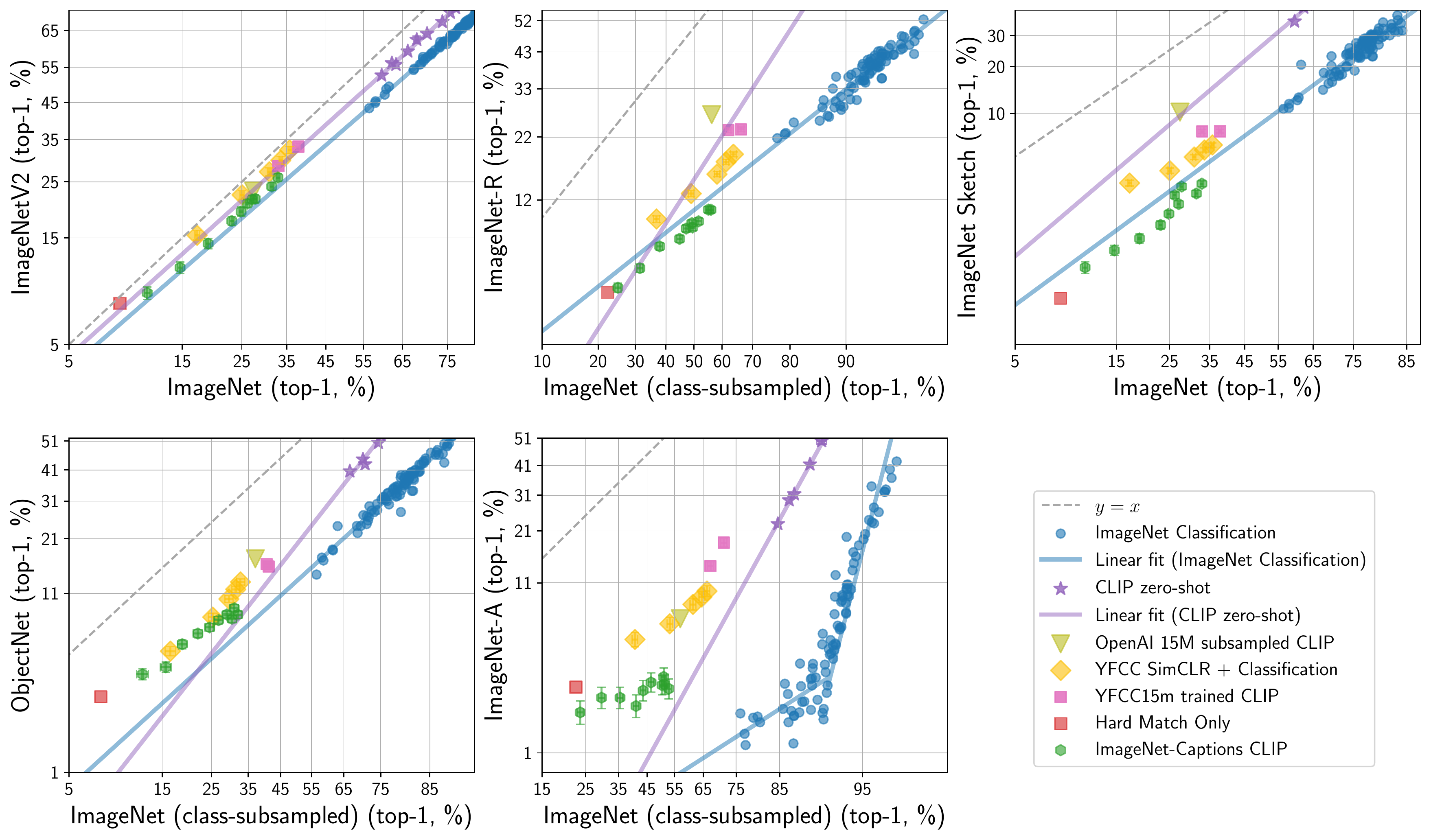}
\end{figure*}

\begin{table*}[hbt!]
\centering
\begin{tabular}{lllllll}
\toprule
\cmidrule(r){1-2}
Experiment  & IN & IN-V2 & IN-R  & IN Sketch & ObjectNet & IN-A  \\
& (top-1, \%) & (top-1, \%) & (top-1, \%) & (top-1, \%) & (top-1, \%) & (top-1, \%) \\
\midrule
YFCC-15M-Cls Only & 8.3 & 7.7 & 4.5 & 0.4 & 2.8 & 2.6 \\
YFCC NoCLIP & 35.7 & 32.4 & 18.8 & 6.0 & 12.7 & 9.9 \\
\bottomrule
\end{tabular}
\vskip -0.1in
\end{table*}

\newpage

\section{ImageNet-Captions additional statistics}
\label{app:imagenet_captions_statistics}

\begin{table}[h]
    \centering
    \rowcolors{1}{}{gray!25}
  \caption{\label{tab:imagenet_captions_languages} The most frequently occurring languages in ImageNet-Captions, according to PYCLD2 top-1. English also appears as a top-3 language in 91\% of the captions.}
  \centering
    \begin{tabular}{lcc}
    \toprule
    Language               & \# Captions & \% of Total \\\midrule
    English                & 416,601   & 89.9 \\
    Chinese                & 5,357   & 1.2 \\
    Spanish                & 3,893   & 0.8 \\
    Danish                & 2,993   & 0.6 \\
    Italian                & 2,598   & 0.6 \\
    German                & 2,263  &   0.5 \\
    Portuguese            & 2,104  &   0.5 \\
    Dutch                & 1,924   &   0.4 \\
    French                & 1,433   &  0.3 \\
    Scottish                & 1,404 &  0.3 \\
    \end{tabular}
\end{table}

\begin{figure*}[h]
    \centering
    \includegraphics[width=0.64\textwidth]{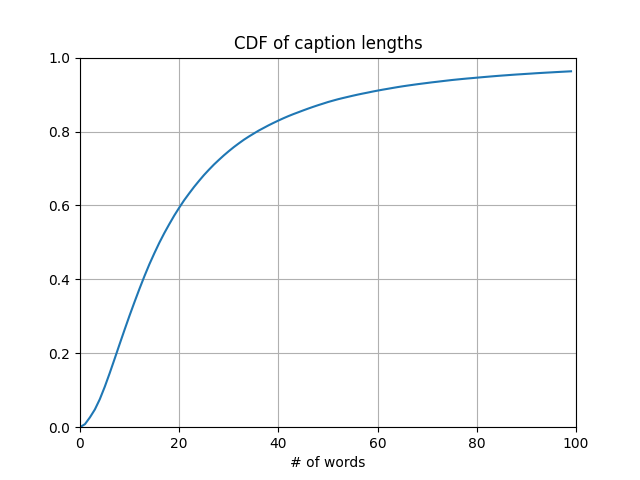}
    \caption{\label{fig:cdf_length} The cumulative distribution function of the caption lengths (in number of words) for the title-tag-description variant of ImageNet-Captions. We limit the maximum of the x-axis to 100 words, as only 3.6\% of captions are between 100 and 4,924 words. The median caption length is 17 words.}
\end{figure*}

\end{document}